\documentclass[lettersize,journal]{IEEEtran}

\IEEEoverridecommandlockouts                              % This command is only needed if 
\usepackage{graphics} % for pdf, bitmapped graphics files

\usepackage{times} % assumes new font selection scheme installed
\usepackage{cite}
\usepackage{multirow}

\usepackage{amsthm}

\usepackage{amssymb}  % assumes amsmath package installed

\usepackage[linesnumbered,ruled,vlined]{algorithm2e}
\SetKwInput{KwInput}{Input}                % Set the Input
\SetKwInput{KwOutput}{Output}              % set the Output

\newcommand{\mR}{\mathbb{R}}
\newcommand{\mE}{\mathbb{E}}
\newcommand{\mP}{\mathbb{P}}
\newcommand{\mQ}{\mathbb{Q}}

\usepackage{subcaption} 
\usepackage{color}

\usepackage{mathtools}
\usepackage{float}

\usepackage{cite}

\usepackage{hyperref}
\hypersetup{
    colorlinks=true,
    linkcolor=blue,
    filecolor=magenta,      
    urlcolor=cyan,
    pdftitle={Overleaf Example},
    pdfpagemode=FullScreen,
    }

\newtheorem{lemma}{Lemma}
\newtheorem{corollary}{Corollary}

\title{
Path Integral Control for Hybrid Dynamical Systems
}

\author{Hongzhe Yu, Diana Frias Franco, Aaron M. Johnson, and Yongxin Chen% <-this % stops a space
\thanks{Financial support from NSF under grants 1942523, 2008513 are greatly acknowledged.}
\thanks{Hongzhe Yu and Yongxin Chen are with the Institute for Robotics and Intelligent Machines, Georgia Institute of Technology, Atlanta, GA; {\{hyu419,yongchen\}@gatech.edu}; 
}
\thanks{Diana Frias Franco and Aaron M. Johnson are with the Robotics Institute, Carnegie Mellon University, Pittsburgh, PA 15213 USA; {\{dfriasfr,amj1\}@andrew.cmu.edu}.}
}

\begin{document}

\maketitle
\thispagestyle{empty}
\pagestyle{empty}

%%%%%%%%%%%%%%%%%%%%%%%%%%%%%%%%%%%%%%%%%%%%%%%%%%%%%%%%%%%%%%%%%%%%%%%%%%%%%%%%
% improve the abstract
\begin{abstract}
This work introduces a novel paradigm for solving optimal control problems for hybrid dynamical systems under uncertainties. Robotic systems having contact with the environment can be modeled as hybrid systems. Controller design for hybrid systems under disturbances is complicated by the discontinuous jump dynamics, mode changes with inconsistent state dimensions, and variations in jumping timing and states caused by noise. We formulate this problem into a stochastic control problem with hybrid transition constraints and propose the Hybrid Path Integral (H-PI) framework to obtain the optimal controller. Despite random mode changes across stochastic path samples, we show that the ratio between hybrid path distributions with varying drift terms remains analogous to the smooth path distributions. We then show that the optimal controller can be obtained by evaluating a path integral with hybrid constraints. Importance sampling for path distributions with hybrid dynamics constraints is introduced to reduce the variance of the path integral evaluation, where we leverage the recently developed Hybrid iterative-Linear-Quadratic-Regulator (H-iLQR) controller to induce a hybrid path distribution proposal with low variance. The proposed method is validated through numerical experiments on various hybrid systems and extensive ablation studies. All the sampling processes are conducted in parallel on a Graphics Processing Unit (GPU).
\end{abstract}

%%%%%%%%%%%%%%%%%%%%%%%%%%%%%%%%%%%%%%%%%%%%%%%%%%%%%%%%%%%%%%%%%%%%%%%%%%%%%%%%
\section{Introduction}
\label{sec:introduction}
% Planning and control under uncertainties
The importance of considering uncertainties in control and robot motion planning is underlined by the ubiquitous existence of noise that arises from modeling and actuation errors, sensor inaccuracies, and external disturbances in robotics systems \cite{thrun2002probabilistic}. Stochastic optimal control \cite{maybeck1982stochastic, aastrom2012introduction} provides a principled formulation to obtain a control signal sequence that minimizes a statistical objective index under system and actuation noise. Extensive efforts have been devoted to obtaining optimal controllers for both linear and nonlinear stochastic systems \cite{bertsekas1996stochastic, kushner1990numerical, kappen2005path, todorov2005generalized, chen2015optimal_1}. However, most existing stochastic control frameworks constrain their formulations within smooth stochastic systems.

% The hybrid systems
The hybrid of continuous and discrete dynamics represents a wide range of scenarios in robotics. Rigid-body dynamics with contact, such as walking \cite{collins2005bipedal, laszlo1996limit, todd2013walking, kuindersma2016optimization}, running \cite{clark2001biomimetic, hutter2011scarleth, westervelt2018feedback}, and manipulation \cite{johnson2016hybrid,billard2019trends} are typical hybrid systems. The \textit{Hybrid Dynamical System}  \cite{grossman1993hybrid} models these phenomena using piecewise smooth vector fields, known as the \textit{flows} connected by discrete logics and reset functions, known as the \textit{jumps}. Nonlinear smooth flows and instantaneous jump dynamics at hybrid events are the main challenges for controlling hybrid systems. The trajectory optimization paradigm formulates optimization problems with constraints on the contacts for hybrid systems, but they often have high time complexity and limited feasible solution space \cite{posa2014direct, mordatch2012discovery, diehl2006fast}. 

Uncertainties affect the stability of hybrid systems through both the smooth flows and the discrete jumps. The studies of the region of attraction (ROA) analysis and control design for limit cycles \cite{manchester2011regions, manchester2011stable, laszlo1996limit} revealed that hybrid systems are intrinsically unstable. Uncertainty control is, therefore, critical for hybrid systems. During a typical execution of a feedback controller that stabilizes the system around a nominal trajectory, or an \textit{`orbit'}, noises and disturbances in the smooth flows lead to randomness in the timing, the mode, and the pre-event states that the actual jumps happen under the designed controller, as opposed to the deterministic nominal plan \cite{yu2024optimal}. Issues like mode mismatch \cite{rijnen2015optimal, rijnen2017control} need careful treatment during the design phase and often lead to suboptimality in the solutions.

\begin{figure}
    \centering
    \includegraphics[width=\linewidth]{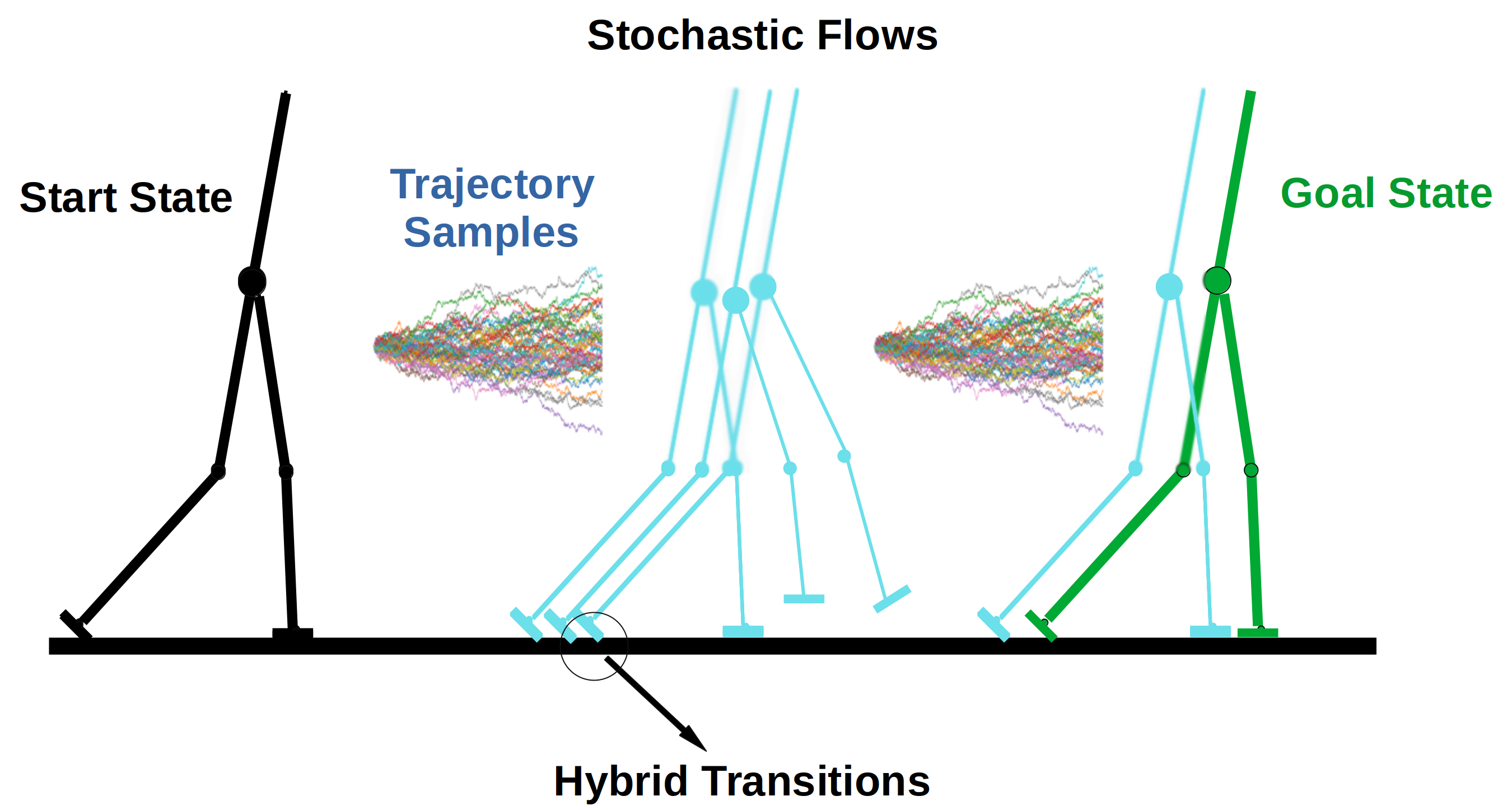}
    \caption{Hybrid Path Integral Sampling Procedure for Hybrid Dynamical Systems with Stochastic Smooth Flows.}
    \label{fig:method_intro}
\end{figure}

One paradigm to approximately solve nonlinear smooth stochastic problems is Differential Dynamic Programming (DDP), or iterative Linear Quadratic Regulator (i-LQR) \cite{li2004iterative, todorov2005generalized}, which iteratively linearizes the nonlinear system and quadratically approximates the cost functions, and solve the approximated problem in closed form. Recently, this paradigm was extended to the hybrid system in \cite{KongHybridiLQR}, where the authors leveraged a first-order approximation of the hybrid jump dynamics using the \textit{Saltation Matrix} (or the \textit{Jump Matrix}) \cite{kong2023saltation} to backpropagate the policy updates through hybrid events, and proposed the Hybrid i-LQR (H-iLQR) strategy. As DDP achieves local optimality for smooth flows, the H-iLQR also relies on the precision of linearization of the guards and resets defined in the Saltation matrix. Furthermore, the feedback nature of the hybrid DDP paradigm faces mode inconsistencies between the stochastic trajectories and the nominals during the jump. Existing methods rely on heuristical designs, which do not guarantee a stabilizing feedback control law. 

Solving general nonlinear smooth stochastic problems globally reduces to solving a nonlinear Hamilton-Jacobian-Bellmen (HJB) equation, which is a partial differential equation (PDE), using Dynamical Programming (DP) \cite{bertsekas2012dynamic}. Adding hybrid guard functions and reset maps in these PDEs makes solving the problem more challenging. For a class of nonlinear smooth problems, the nonlinear HJB becomes linear under a log transform of the cost-to-go \cite{kappen2005linear}, and the optimal control can be obtained by evaluating a path integral over a path distribution induced by Stochastic Differential Equations (SDEs) \cite{kappen2005path}. The path integral control paradigm mitigates the scalability issues of DP by leveraging parallel computation of the path integral. More importantly, for hybrid systems, the jump dynamics are not linearized or approximated; only direct detection of the guard conditions and applications of the resets are required. This paradigm eliminates the error induced by smooth and Saltation linearizations. Finally, different from feedback controllers, the forward-sampling-based optimal controller does not compute the state deviations from a nominal trajectory and thus circumvents the problem of mode mismatches. Path integral control has been applied in robotics tasks \cite{GradyWilliamsICRA, GradyWilliamsTRO} that consider smooth dynamics where the robotic systems do not have physical interactions with the environment, limiting its applications in a range of critical robotics applications \cite{grizzle2014models, westervelt2018feedback, kuindersma2016optimization}.

In this work, we solve stochastic control problems for hybrid systems with optimality by formulating the problem into a stochastic optimal control problem with hybrid guard and reset constraints. We consider smooth nonlinear stochastic dynamics connected by deterministic hybrid transitions. These describe the situations where deterministic mechanisms determine the triggering conditions between the modes, and the smooth flows are subject to disturbances. The uncertainties affect the times and the states at which the hybrid transitions happen for a rollout but not the underlying mechanism. The stochastic smooth flows with hybrid transitions form a hybrid path distribution under a given controller. We then convert the stochastic control problem into a path distribution control problem, where the optimal controller is expressed as a path integral over SDEs with hybrid transitions. 

One potential issue of path integral methods is the low sampling efficiency caused by the high variance of the underlying distribution from which we sample. Importance sampling is often leveraged in the path distribution space to reduce the variance \cite{zhangpath, zhang2023optimal}. In this work, we leverage the generalized importance sampling in the hybrid path distribution space and use the H-iLQR controller as a proposal controller in the hybrid path distribution space. 

\subsection{Related Works}
{\em (a) Smooth Stochastic Control and Planning.}
The nonlinear HJB can be converted to a linear PDE for a particular class of nonlinear stochastic systems \cite{kappen2005path, kappen2007introduction, kappen2005linear}. The solution can then be transformed into a path integral by leveraging the Feynman-Kac formula \cite{del2004feynman}, and the optimal controller can be obtained by efficiently sampling SDEs. The smooth path integral control paradigm can be viewed as a policy improvement technique \cite{theodorou2010generalized}. In the linear quadratic case, the HJB is reduced to the Riccati equation, which can be solved in closed-form with additional constraints \cite{chen2015optimal_1, chen2015optimal_2, yongxin2018optimal_3}. Smooth stochastic control is dual to path distribution control and inference \cite{kappen2012optimal, rawlik2013stochastic} and cross-entropy methods \cite{zhang2014applications}, where the search for the optimal controller is converted into the search for an optimal controlled path distribution. This paradigm finds applications in motion planning tasks \cite{yu2023gaussian, yu2023stochastic, power2024constrained}.

{\em (b) Optimal Control for Robotics Systems with Contacts.}
Rigid contacts with frictions can be modeled as linear or non-linear complementarity constraints (LCPs or NCPs) in an optimization program in \cite{mangasarian1993nonlinear, hager2009nonlinear}. Leveraging this modeling, contact-implicit trajectory optimization \cite{posa2014direct, manchester2019contact, patel2019contact} solves an optimal control problem with LCP or NCP constraints, which gives a control sequence that both optimizes the objective index and satisfies the hybrid dynamics constraint with contacts. More recently, in \cite{KongHybridiLQR}, the DDP / iLQR paradigm was extended to hybrid systems by leveraging the Saltation Matrix approximation of the hybrid transitions \cite{filippov2013differential, munoz2019enhancing, kong2023saltation}, and the hybrid events are directly detected by the forward passes in the hybrid DDP. This paradigm is more efficient and was applied in the model predictive control for hybrid systems.

{\em (c) Robust Control for Hybrid Systems under Uncertainties.} 
Considering uncertainties on top of the intrinsic instabilities of a hybrid system, designing a robust motion plan or controller under uncertainty becomes a natural quest for contact-rich systems \cite{belter2011rough, tassa2011stochastic, dai2012optimizing,shirai2024chance}. In \cite{dai2012optimizing}, the authors minimize the cost-to-go function to obtain a robust limit cycle for legged robots. In \cite{tassa2011stochastic, drnach2021robust}, the authors considered uncertain contacts and proposed a formulation that minimizes the expected residual of the stochastic contact constraints. 

\subsection{Contributions}

In this work, we propose the \textit{Hybrid Path Integral} (H-PI) framework, which leverages the duality between the stochastic control and path distribution control problems to obtain the optimal controller for stochastic systems subject to hybrid transition conditions. Unlike in the smooth case, each stochastic path sample is subject to random changes in the state space, dimension, and dynamics, collectively decided by the stochastic smooth flows, guard conditions, and reset functions. Despite these differences, we show that the optimal controller can be obtained via efficient parallelizable forward sampling of SDEs under hybrid constraints.

The contributions of this work are summarized as follows:
\begin{enumerate}
    \item We show that Girsanov's theorem on the change of path measures induced by SDEs with hybrid transitions remains similar to the smooth stochastic processes. 

    \item We show that stochastic control problems with hybrid transitions can be transformed into a hybrid distribution control, and the optimal controller is obtained by evaluating a path integral under stochastic trajectories with hybrid transitions. 

    \item We propose an efficient parallelizable algorithm to sample future stochastic trajectories subject to hybrid transitions, where we use the H-iLQR controller to guide the sampling process for lower sampling variance. 
    
    \item Theoretical proofs, extensive experiments, and ablation studies are conducted to validate the proposed paradigm. 
    
\end{enumerate}

The rest of the paper is organized as follows: Section \ref{sec:preliminaries} introduces essential preliminaries, Section \ref{sec:problem_formulation} defines the problem formulation, Section \ref{sec:hybrid_pi_theory} derives the optimal controller, Section \ref{sec:importance_sampling} presents the importance sampling scheme for hybrid path distributions, Section \ref{sec:method} formalizes the main algorithm, and Section \ref{sec:experiments} presents experiment results, concluded by the Section \ref{sec:conclusion}.

\section{Preliminaries}\label{sec:preliminaries}
This section introduces preliminary knowledge, including hybrid dynamical systems and smooth path integral control.

\subsection{Smooth Path Integral Control}
\label{sec:preliminaries_smooth_PI}
Smooth stochastic control problem considers the following nonlinear stochastic dynamical system 
\begin{equation}
\label{eq:controlled_nonlinear_SDE_smooth}
d X_t = F(t, X_t) d t + \sigma(t,X_t) (u_t d t + \sqrt{\epsilon} d W_t)
\end{equation}
in the time window $t\in[0, T]$, and aims to minimize the following objective \cite{kappen2005path, thijssen2015path, zhang2023optimal}
\begin{equation}
\label{eq:expected_cost_smooth}
\begin{split}
  \mathcal{J} \coloneqq  &\; \mathbb{E}  \left[ \int_0^T \left( V(t,X_t) + \frac{1}{2} \lVert u_t\rVert^2 \right) dt + \Psi_T(X_T) \right].
\end{split}
\end{equation}
Here $\sigma$ is the mapping from control to states, and $dW_t$ is a standard Wiener process with intensity $\epsilon$ coming into the system via the same channel as the control. $V(t,X_t)$ and $\Psi_T(X_T)$ are a running and a terminal cost, respectively. DP leads to solving the HJB PDE
\begin{align*}
% \label{eq:nonlinearHJB}
\!\!\! - \partial_t \mathcal{J}_t
=
\min_u \left(V + \partial_x \mathcal{J}_t \left( F + \sigma u \right)
+ \frac{\epsilon}{2} \mathrm{Tr} \left( \sigma \sigma' \partial_{xx} \mathcal{J}_t \right) \right),
\end{align*}
where $\mathcal{J}_t$ is the cost-to-go \eqref{eq:expected_cost_smooth} in time window $[t,T]$.
The change of variable $\psi \coloneqq \exp \left( - \frac{1}{\epsilon} \mathcal{J} \right)$ transforms the above nonlinear PDE into a linear one \cite{kappen2005path}
\begin{equation}
\label{eq:linearHJB}
   \partial_t \psi + \partial_x \psi F + \frac{\epsilon}{2} \mathrm{Tr} ( \sigma \sigma' \partial_{xx} \psi ) = \frac{1}{\epsilon}V \psi 
\end{equation}
with terminal condition $\psi_T = \exp(-\frac{1}{\epsilon}\Psi_T)$. The optimal control is obtained in a state feedback form $u_t^* = -\sigma' \partial_x\psi(t, x_t)$. Leveraging Feynman Kac's formula \cite{del2004feynman}, this optimal controller can be written as a path integral over the path induced by SDEs.

{\em Importance sampling for smooth path distributions.}
Importance sampling increases the efficiency of evaluating an expectation by change of measures. Consider the probability measure $\mathbb{P}^0$ induced by the uncontrolled SDE
\begin{equation}
\label{eq:nonlinear_SDE_uncontrolled}
d X_t = F(t, X_t) d t + \sqrt{\epsilon} \sigma(t,X_t) d W_t.
\end{equation}
The ratio between the measure $\mP^u$ induced by the process \eqref{eq:controlled_nonlinear_SDE_smooth}, and the measure $\mP^0$ satisfy \cite{Gir60} 
\begin{equation}
\label{eq:Girsanov_SDE_measure_ration}
    \frac{d \mP^u}{d \mP^0} = \exp \left (\int_0^T \frac{1}{2\epsilon} \lVert u_t \rVert^2 d t + \frac{1}{\sqrt{\epsilon}} u_t'dW_t \right).
\end{equation}
The optimal control that minimizes \eqref{eq:expected_cost_smooth} is expressed as the following path integral over the controlled stochastic path distribution \cite{kappen2005path, thijssen2015path} as 
\begin{equation}
    u^*_0 = u_0 + \lim_{\Delta t \to 0}\frac{\mE_{\mP^u} \left[ \sqrt{\epsilon}\int_t^{t+\Delta t} \exp(-\frac{1}{\epsilon}S^u(t)) dW_s \right] }{\Delta t \times \mE_{\mP^u} \left [ \exp(-\frac{1}{\epsilon}S^u(t)) \right]},
\end{equation}
where $u_t$ is a proposal control at time $t$, and
\begin{equation}
\label{eq:Su_definition}
    \!\!\!\! S^u(t) \triangleq \!\! \int_{t}^{T} V(X_\tau) + \frac{1}{2} \lVert u_\tau \rVert^2 d\tau + \sqrt{\epsilon} u_\tau'dW_\tau  + \Psi_T(X_T).
\end{equation}
\subsection{Hybrid Systems and Jump Dynamics Linearization}
\label{sec:hybrid-ilqr}
{\em (a) Hybrid Dynamical Systems.}
A hybrid dynamical system contains modes with smooth dynamics and transitions between them. A hybrid system is defined by a tuple \cite{grossman1993hybrid, kong2023saltation} $\mathcal{H} \coloneqq \{ \mathcal{I}, \mathcal{D}, \mathcal{F}, \mathcal{G}, \mathcal{R} \}.$
The set 
\begin{equation}
    \mathcal{I} \coloneqq \{ I_1, I_2, \dots, I_{N_I}\} \subset \mathbb{N}
\label{eq:hybrid_definitions}
\end{equation}
is a finite set of \textit{modes}, $\mathcal{D}$ is the set of continuous \textit{domains} representing the state spaces, with individual domains $D_{j}$ for each mode $I_j$, $\mathcal{F}$ is the set of \textit{flows}, consisting of individual flows $F_{j}$ describes the smooth dynamics in mode $I_j$. $\mathcal{G} \subseteq \mathcal{D}$ denotes the set of \textit{guards} triggering the transitions. A transition from mode $I_j$ to mode $I_k$ happens at state $X_t^-:=X(t^-)$ and time $t^{-}$ if 
\begin{equation}
\label{eq:guard_func_j}
    g_{jk}(t^{-}, X_t^{-}) \leq 0, \; X_t^{-} \in I_j.
\end{equation}
The jump dynamics are defined as
\begin{equation}
\label{eq:reset_map_j}
    X_t^{+} = R_{jk}(X_t^{-}) \in I_k,
\end{equation}
where $R_{jk} \in \mathcal{R}$ is the corresponding \textit{reset function} mapping state from $D_{j}$ to $D_{k}$ when the guard condition $G_{jk}$ is met. We use $(t^-, X_t^-)$ and $(t^+, X_t^+)$ to represent the pre-impact and post-impact time-state pairs. 

{\em (b) Jump Dynamics Linearization and Hybrid iLQR.}
The reset dynamics $X_t^+ = R(t, X_t^-)$ is generally instantaneous and nonlinear. The Saltation Matrix \cite{aizerman1958stability, filippov2013differential, munoz2019enhancing, kong2023saltation} is a first-order approximation for the varying effect of the transition around a nominal point. From mode $I_j$ to $I_k$, it is defined as 
\begin{equation}
\label{eq:saltation}
    \Xi_{jk} \triangleq \partial_x R_{jk} + \frac{(F_{k} - \partial_x R_{jk} \cdot F_{j} - \partial_t R_{jk})\partial_x g_{jk}}{\partial_t g_{jk} + \partial_x g_{jk} \cdot F_{j}},
\end{equation}
where $\partial_x(\cdot)$ and $\partial_t(\cdot)$ denote the partial derivatives in state and time, respectively. The dynamics of a perturbed state $\delta X_t$ at hybrid transitions can be approximated as
\begin{equation}
\label{eq:saltation_approximation_dyn}
    \delta X_t^+ \approx \Xi_{jk} \delta X_t^-.
\end{equation}
The Saltation approximation performs better than directly differentiating the reset map by considering time differences in hitting the guard \cite{kong2023saltation}.
Based on this linearization technique, the Hybrid iterative Linear-Quadratic-Regulator (H-iLQR) was introduced in \cite{kong2021ilqr, KongHybridiLQR}.

\section{Problem Formulation}
\label{sec:problem_formulation}
This section introduces the main problem formulation. 

{\em Mode-dependent Notations.} The quantities and matrices that will appear hereafter, i.e., ($n,m, X_t,u_t,F(t,X_t),\sigma,W_t$), etc, all have the same definitions as in the smooth problem in Section \ref{sec:preliminaries_smooth_PI} with a mode dependent notation $j$ to indicate the mode $I_j$ they are in.

\subsection{Stochastic Systems with Hybrid Transitions}
\label{sec:problem_formulation_SOCproblem}

For the hybrid system \eqref{eq:hybrid_definitions}, \eqref{eq:guard_func_j}, \eqref{eq:reset_map_j}, we consider mode-dependent state $X^j_t \in \mR^{n_j}$ and control $u^j_t \in \mR^{m_j}$ variables in mode $I_j$, where the \textit{controlled} smooth flow is written as 
\begin{equation}
\begin{split}
    \label{eq:smooth_dyn_mode_j}
     \!\!\!\!\! d X^j_t = F_{j}(t, X^j_t) d t + \sigma_{j}(t, X^j_t) (u^j(t, X^j_t) d t + \sqrt{\epsilon} d W^j_t). 
\end{split}
\end{equation}
We define the \textit{uncontrolled} smooth flow in mode $I_j$ as
\begin{equation}
\begin{split}
    \label{eq:uncontrolled_smooth_dyn_mode_j}
    \!\!d X^j_t =& F_{j}(t, X^j_t) d t + \sqrt{\epsilon} \sigma_{j}(t, X^j_t) d W^j_t.
\end{split}
\end{equation}
The jump dynamics are defined by guard \eqref{eq:guard_func_j} and reset \eqref{eq:reset_map_j}. 

For a given control sequence $u(t,X_t)$ and a realization of the random process $dW_t$, the state trajectory rollout $\{X_t | t \in [0,T]\}$ is assumed to have $N_J$ hybrid transitions at time $\{t^-_j\}_{j=1,\dots,N_J}$. We extend the jump time set by letting $t_0^+ = 0$ and $t_{N_J+1}^{-}=T$. For simplicity, we assume that the system is in mode $I_j$ in the time $[t_j^+, t_{j+1}^-]$, i.e., 
\begin{align*}
    X_t^j \in I_j, \forall t \in [t_j^+, t_{j+1}^-]; \; X^{j+1}(t^+_{j+1}) = R_{j, j+1}(X^{j}(t^-_{j})).
\end{align*}

\subsection{Stochastic Optimal Control with Hybrid Transitions}
The stochastic optimal control problem we consider is defined in the time window $[0,T] = \cup_{j=0,\dots,N_J}[t_j^+, t_{j+1}^-]$ as 
\begin{align}
\label{eq:formulation_main}
  \min_u & \; \mathcal{J}_H \triangleq \mathbb{E} \left[ \sum_{j=0}^{N_J}  \int_{t^+_j}^{t^-_{j+1}} \left( V(t,X^j_t) + \frac{1}{2} \lVert u^j_t\rVert^2 \right) dt + \Psi_T \right]
  \nonumber
  \\
  {\rm s.t.} \; & \eqref{eq:smooth_dyn_mode_j}, \; \eqref{eq:guard_func_j}, \; \eqref{eq:reset_map_j}, \; \forall j=1,\dots, N_I.
\end{align}

{\em Time Discretizations.} As time discretization is unavoidable in numerical simulations, we consider a length-$N_T$ time discretization over the window $[0, T]$ defined as
\begin{equation}
\label{eq:time_discretization}
    t_1=0; \; t_{i+1} = t_i + \Delta t, i=1,\dots, N_T-1; \; t_{N_T} = T.
\end{equation}
where $\Delta t = T/N_T$. 

Let $F^H_{\Delta,i}(X^j_i, u_i)$ represent the hybrid transition from $X^j_i$ to $X_{i+1}$. If the guard function $g_{j,j+1}(t_{i+1}, X_{i+1}) > 0$, then $X^j_{i+1} \in I_j$ is still in $I_j$, and
\begin{equation}
\begin{split}
    X^j_{i+1} = &F^H_{\Delta, i}(X^j_i, u^j_i) \approx  X^j_i + (F_j(t_i, X^j_{t_i}) + 
    \\
    &\sigma_j(X_t^j) u^j_{t_i})) \Delta t + \sqrt{\epsilon} \sigma_j(X_t^j) \Delta W^j_i.
\end{split}
\label{eq:discretetime_smooth_map}
\end{equation}
where $\sqrt{\epsilon}\Delta W_i$ is a discrete increment of Wiener process with intensity $\epsilon$, i.e., $\sqrt{\epsilon}\Delta W_i \sim \mathcal{N}(0, \epsilon \Delta t I)$
Alternatively, if there is a jump at $t^{-}_{j} \in [t_i, t_{i+1}]$, we modify the next time stamp to be $t_{i+1} = t^{-}_{j}$ and $\Delta \Tilde{t}_i = t^{-}_{j} - t_i$. The one-step increment of the Wiener process also changes to 
\[
\sqrt{\epsilon} \Delta \Tilde{W}^j(t_i,t_j^-) \sim \mathcal{N}(0, \epsilon \Delta \Tilde{t}_i I).
\] 
The hybrid dynamics with a jump are 
\begin{equation}
\label{eq:discretetime_jump_map}
\begin{split}
     & X^{j+1}_{i+1} = F^H_{\Delta, i}(X^j_i, u^j_i) 
     \\
    \!\!\! \approx & R_{j,j+1} ( X^j_i + F_j(t_i, X^j_{t_i}) + \sigma_j(X^j_{t_i}) u^j_{i}) \Delta \Tilde{t} + \Delta \Tilde{W}^j).
\end{split}
\end{equation}
The event time $t_j^-$ can be found using bi-section root-finding methods for the guard function inside $[t_i, t_{i+1}]$.

\section{Path Integral Control for Hybrid Systems}
\label{sec:hybrid_pi_theory}

In this section, we establish the optimal control for the problem \eqref{eq:formulation_main} via path distribution control for hybrid systems. 

\subsection{Ratio Between Hybrid Path Distribution Measures}
\label{sec:hybrid_Girsanov}
The Wiener process $W_t$ added to the smooth flow \eqref{eq:smooth_dyn_mode_j} in each mode and the hybrid transitions \eqref{eq:reset_map_j} under the guard conditions \eqref{eq:guard_func_j} lead to a \textit{hybrid path distribution}. 

Consider the following two smooth processes in mode $I_j$
\begin{subequations} 
\label{eq:nonlinear_SDE}
\begin{align}
    dX^j_t &= F_{j}(t, X^j_t) d t + \sqrt{\epsilon} \sigma_j(t, X^j_t) d W^j_t, 
    \label{eq:nonlinear_SDE_1}
    \\
    dX^j_t &= H_{j}(t, X^j_t) d t + \sqrt{\epsilon} \sigma_j(t, X^j_t) d W^j_t, 
    \label{eq:nonlinear_SDE_2}
    \end{align}
\end{subequations}
with hybrid events defined by \eqref{eq:guard_func_j} and \eqref{eq:reset_map_j}. We use $\mP_H$ to denote the measure induced by stochastic dynamics \eqref{eq:nonlinear_SDE_1} and jumps \eqref{eq:reset_map_j}, and let $\mQ_H$ denote the measure induced by \eqref{eq:nonlinear_SDE_2}, \eqref{eq:reset_map_j}. We formally define the two measures as in the smooth case \cite{sarkka2019applied}.

For the time discretization \eqref{eq:time_discretization}, we consider a realization of the Wiener process $\{\Delta W^j_k\}_{N_T}$ under $\mP_H$. Without the loss of generality, the stochastic trajectory sample $\{X^j_k\}_{N_T}$ is assumed to have only \textit{one} jump at $k^{-} \in \{0, \dots, N_T\}$, resetting the state from mode $I_1$ to mode $I_2$. i.e., $X^1_k \in I_1, \forall k<k^{-}$ and $X^1_k \in I_2, \forall k\geq k^{-}$. The measure $d \mP_H$ associated with $\{X^j_k\}$ is the joint probability
\begin{equation}
\label{eq:path_measure_mP}
    d \mP_H = \prod_{k=0}^{k^{-}-1} p(X^1_{k+1}|X^1_k) \prod_{k=k^{-}}^{N_T} p(X^2_{k+1}|X^2_k).
\end{equation}
For the same trajectory rollout, the measure induced by hybrid system \eqref{eq:nonlinear_SDE_1}, \eqref{eq:reset_map_j} is the joint probability
\begin{equation}
\label{eq:path_measure_mQ}
    d \mQ_H = \prod_{k=0}^{k^{-}-1} q(X^1_{k+1}|X^1_k) \prod_{k=k^-}^{N_T} q(X^2_{k+1}|X^2_k).
\end{equation}
Here, the trajectory differs from the smooth dynamics case in that two consecutive states may have different physical meanings due to reset maps. Nevertheless, the critical observation is that the probability measure is independent of the sampled trajectories. Next, we derive the Girsanov theorem for changing measures in the hybrid path distribution space in the following Lemma \ref{thm:change_of_measure}.
\begin{figure}
    \centering
    \includegraphics[width=0.95\linewidth]{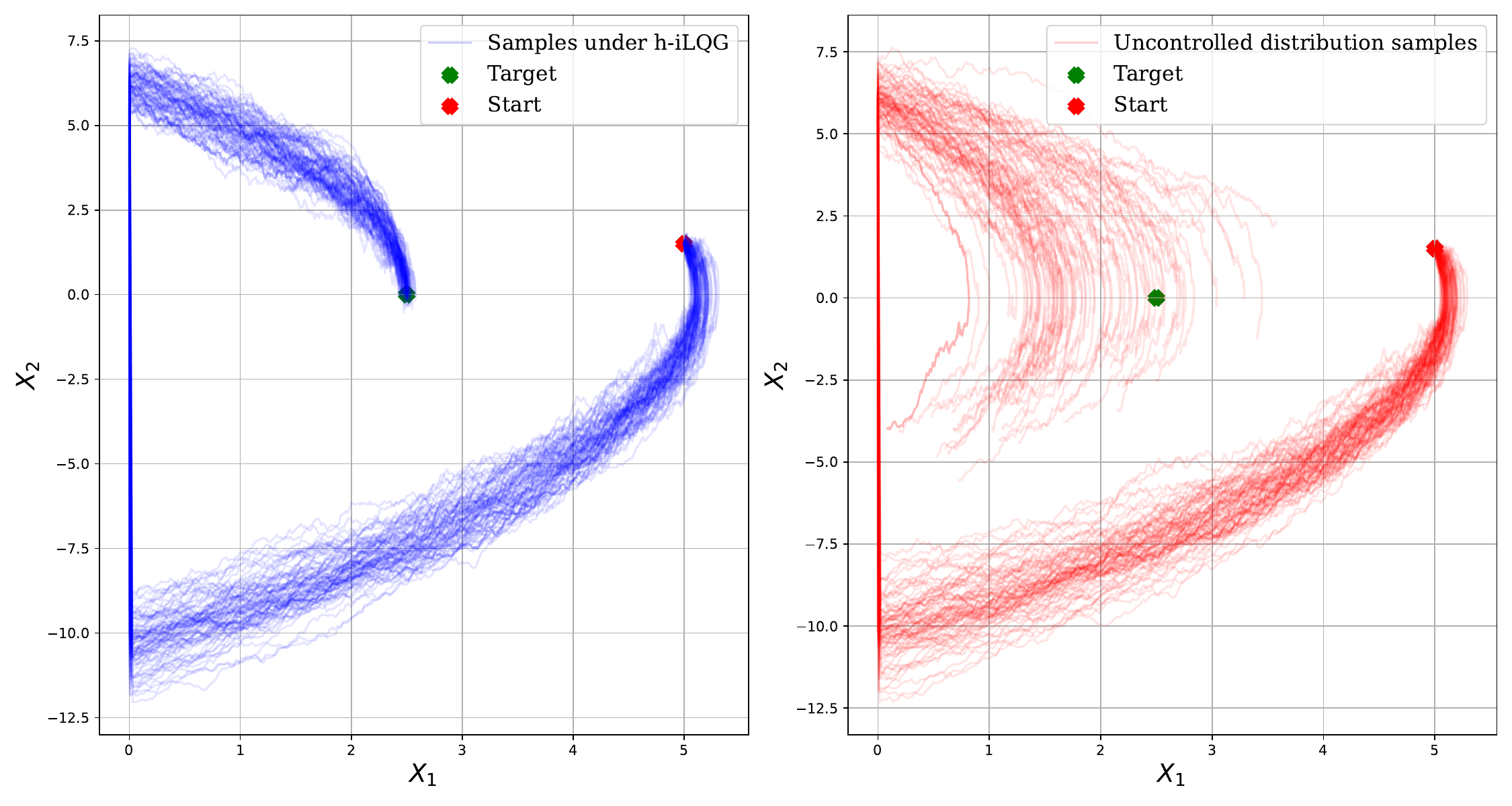}
    \caption{Samples from the controlled (left) and uncontrolled (right) system state distributions for the bouncing ball dynamics. Lemma \ref{thm:change_of_measure} states that the KL-divergence between the two distributions equals the expected control energy.}
    \label{fig:girsanov}
\end{figure}

\begin{lemma}
\label{thm:change_of_measure}
    The ratio between the measures $d \mQ_H$ and $d \mP_H$ is given by 
    \begin{equation*}
    \begin{split}
        \!\!\!\frac{d\mQ_H}{d\mP_H} = \exp \left( \sum_{j=0}^{N_J}  \int_{t^+_j}^{t^-_{j+1}} \frac{(H_{j}-F_{j})' (\sigma_j \sigma_j')^{-1} d W^j_\tau}{\sqrt{\epsilon}}  \right.
        \\
        \!\!\! \left. - \frac{\lVert H_{j}-F_{j} \rVert_{(\sigma_j \sigma_j')^{-1}}^2}{2\epsilon} d \tau \right),
    \end{split}
    \end{equation*}
    where $\{(t_j^{-}, t_j^{+})\}_{N_J}, t_{N_J+1}^{-} = T$ are the jump times of a rollout of hybrid system \eqref{eq:nonlinear_SDE_1}, \eqref{eq:reset_map_j} under measure $d\mP_H$.
\begin{proof}
    See Appendix-\ref{sec:appendix_proof_change_of_measure}.
\end{proof}
\end{lemma}
\begin{corollary}
\label{thm:Girsanov}
The ratio between the measure $d\mP^u$ induced by the controlled process \eqref{eq:smooth_dyn_mode_j}, and the measure $d \mP^0$ induced by the uncontrolled process \eqref{eq:uncontrolled_smooth_dyn_mode_j} with the same jumps conditions \eqref{eq:reset_map_j}, is given by 
\begin{subequations}
\begin{align}
        \!\!\!\!\!\frac{d\mP^u}{d \mP^0} &= \exp \left( \sum_{j=0}^{N_J}  \int_{t^+_j}^{t^-_{j+1}} -\frac{ \lVert u^j_t \rVert^2}{2\epsilon} d t + \frac{1}{\sqrt{\epsilon}} (u^j_t)' d W^j_t  \right)
        \label{eq:ratio_dPu_dP0_1}
        \\
        &= \exp \left( \sum_{j=0}^{N_J}  \int_{t^+_j}^{t^-_{j+1}} \frac{1}{2\epsilon} \lVert u^j_t \rVert^2 d t + \frac{1}{\sqrt{\epsilon}} (u^j_t)' d \Tilde{W}^j_t  \right)
        \label{eq:ratio_dPu_dP0_2}
        \\
        &\coloneqq\exp\left( \frac{1}{\epsilon} \Lambda_H \right),
        \label{eq:ratio_dPu_dP0_3}
\end{align}
\end{subequations}
where $dW_t$ is a Wiener process under $\mP^0$, and $\Tilde{W}_t$ is a Wiener process under $\mP^u$.
\end{corollary}
\subsection{Hybrid Path Distribution Control}
\label{sec:hybrid_pi}
This section formulates the stochastic control formulation \eqref{eq:formulation_main} into a distributional control subject to hybrid transitions. We consider a hybrid sample trajectory $\{X_t\}_{t=0}^{T}$ under a controller $u_t$, and define the piecewise state cost as
\begin{align}
\label{eq:Lu_defn}
    \mathcal{L}_H(t) \triangleq 
    \mathcal{I}_L[t, t^-_{j_m-1}]
    + \sum_{j=j_m}^{N_J} \mathcal{I}_L[t^+_j, t^-_{j+1}],
\end{align}
where $j_m$ is the minimum $j$ that $t^{+}_{j} \geq t$, and $\mathcal{I}_L$ is defined as
\begin{align}
\begin{split}
    \mathcal{I}_L[a, b] \triangleq \int_{a}^{b}  V(\tau,X^j_\tau) d\tau + \Psi_T(X^{N_J}_T).  
\end{split}
\end{align} 
Taking the expectation over $\mP^u$ to the log probability density ratio between the controlled and the uncontrolled hybrid processes \eqref{eq:ratio_dPu_dP0_2}, the objective in \eqref{eq:formulation_main} can be re-formulated as a piecewise distributional control problem \cite{yu2024optimal}
\begin{subequations}
\label{eq:formulation_distributional}
    \begin{eqnarray}
        \!\!\!\!\!\!\!\!\mathcal{J} \!\!\!\!\!&=&\!\!\!\!\! \mathbb{E}_{\mP^u} \! \left[ \sum_{j=0}^{N_J}  \int_{t^+_j}^{t^-_{j+1}} \left( V(t,X^j_t) + \frac{1}{2} \lVert u^j_t\rVert^2 \right) dt + \Psi_T \right]
        \\
        \!\!\!\!\!\!\!\!\!\!\!&=&\!\!\!\!\! \left\langle \mathcal{L}_H, d\mP^u \right\rangle + \mE_{\mP^u} \! \left[ \epsilon \log \frac{d\mP^u}{d\mP^0} \right],
    \end{eqnarray}
\end{subequations}
where $\langle \cdot, \cdot \rangle$ represents the inner product. Formulation \eqref{eq:formulation_distributional} is convex in variable distribution $\mP^u$, and $\mP^u$ is subject to the constraint $\langle 1, \mP^u \rangle = 1$. Introduce a multiplier $\xi$ for this constraint; the Lagrangian of the problem is
\[
L_{\mathcal{J}} = \langle \mathcal{L}_H + \epsilon \log \mP^u - \epsilon \log\mP^0 + \xi , \mP^u \rangle - \xi. 
\]
The necessary condition for the optimal distribution $d\mP^*$ under the optimal control $u^*$ is that the first-order variation 
\[
0 = \delta L_{\mathcal{J}} |_{\mP^*} = \mathcal{L}_H + \epsilon \left(\log \mP^* - \log\mP^0\right) + \xi + 1,
\]
leading to $d\mP^* \propto d\mP^0 \exp \left( - \frac{1}{\epsilon}\mathcal{L}_H \right)$. After normalization,
\begin{equation}
\label{eq:optimal_distribution_P_star}
    \!\!\!d\mP^* = \frac{\exp \left( -\frac{1}{\epsilon}\mathcal{L}_H \right) d\mP^0  }{\int \exp \left( -\frac{1}{\epsilon}\mathcal{L}_H \right) d\mP^0 } = \frac{\exp \left( -\frac{1}{\epsilon}\mathcal{L}_H \right)}{ \mE_{\mP^0} \left[ \exp \left( -\frac{1}{\epsilon}\mathcal{L}_H \right) \right]} d\mP^0.
\end{equation}

\subsection{Optimal Controller as Hybrid Path Integral}
The search for the optimal controller is obtained by minimizing the KL-divergence between the optimal distribution $\mP^*$ and a controlled distribution $\mP^u$. This formulation is inspired by the cross-entropy methods for the optimal control for smooth stochastic systems \cite{zhang2014applications} and was first introduced to the robotics community for applications in autonomous driving \cite{GradyWilliamsICRA, GradyWilliamsTRO}. The cross-entropy formulation is
    \begin{align}
        \label{eq:KL_P_optimal_Pu}
        \begin{split}
            \!\!\!\!\!\!\!\! u^* &= \arg\min_u {\rm KL} \left( \mP^* \parallel \mP^u \right) = \arg\min_u \mE_{\mP^*} \left[ \log \frac{d \mP^*}{d \mP^0} \frac{d \mP^0}{d \mP^u} \right]
            \\
            &= \arg\min_u \mE_{\mP^*} \left[ \log \frac{\exp \left( - \frac{1}{\epsilon} \left( \mathcal{L}_H + \Lambda_H \right) \right)}{ \mE_{\mP^0} \left[ \exp \left( -\frac{1}{\epsilon} \mathcal{L}_H \right) \right]} \right]
            \\
            &= \arg\min_u \mE_{\mP^*} \left [ -\frac{1}{\epsilon} \Lambda_H \right ],
        \end{split}
    \end{align}
    since the term $\mathcal{L}_H$ is independent to the control. $\Lambda_H$ has different forms \eqref{eq:ratio_dPu_dP0_1} and \eqref{eq:ratio_dPu_dP0_2} when considering Wiener processes under different measures. Minimizing the above over the control $u_t$ gives the optimal controller, introduced in the following Lemma \ref{thm:main_optimal_control}.

\begin{lemma}
    \label{thm:main_optimal_control}
    The optimal controller for the problem \eqref{eq:KL_P_optimal_Pu} at time $t$ can be expressed as an expectation 
\begin{equation}
\label{eq:ustar_P0}
u_t^* = \frac{\sqrt{\epsilon} \times \mE_{\mP^0} \left[ \Delta W_t \exp \left( -\frac{1}{\epsilon}\mathcal{L}_H(t) \right) \right] }{\Delta t \times \mE_{\mP^0} \left[ \exp \left( -\frac{1}{\epsilon}\mathcal{L}_H(t) \right) \right] }.
\end{equation}
\begin{proof}
    See Appendix \ref{sec:proof_optimal_control_expectation}.
\end{proof}
\end{lemma}
The optimal control law is expressed as an expectation over the uncontrolled processes, $\mP^0$, of a function consisting of the path costs collected along the stochastic trajectories with hybrid transitions. 

\section{Hybrid Path Distribution Importance Sampling}
\label{sec:importance_sampling}
The efficiency of evaluating the expectation \eqref{eq:ustar_P0} depends on the variance of the estimated random variable under $\mP^0$. Fig. \ref{fig:girsanov} illustrates an example with only a terminal cost. The uncontrolled process has a high variance in the terminal costs compared with the H-iLQR-controlled process.

\subsection{Importance Sampling in Hybrid Path Distribution Space}
Importance sampling in the hybrid path distribution space is leveraged to reduce the variance of \eqref{eq:ustar_P0} by replacing the underlying measure $\mP^0$ with another path measure $\mP^u$. In this work, we choose $\mP^u$ as the one induced by the H-iLQR. Given the ratio \eqref{eq:ratio_dPu_dP0_2}, we have 
\begin{equation*}
   \mE_{\mP^0} \left[ \Delta W_t \exp \left( -\frac{\mathcal{L}_H}{\epsilon} \right) \right] = \mE_{\mP^u} \left[ \Delta W_t \exp \left( -\frac{\mathcal{S}_H}{\epsilon} \right) \right],  
\end{equation*}
and 
\[
\mE_{\mP^0} \left[ \exp \left( -\frac{\mathcal{L}_H}{\epsilon} \right) \right] = \mE_{\mP^u} \left[ \exp \left( -\frac{\mathcal{S}_H}{\epsilon} \right) \right],
\]
where $\mathcal{S}_H^u(t)$ is defined as the piece-wise smooth integration
\begin{align}
\label{eq:Su_defn}
    \mathcal{S}_H^u(t) \triangleq 
    \mathcal{I}_S[t, t^-_{j_m-1}]
    + \sum_{j=j_m}^{N_J} \mathcal{I}_S[t^+_j, t^-_{j+1}],
\end{align}
where $j_m$ is the minimum $j$ that $t^{+}_{j} \geq t$, and each integral $\mathcal{I}_S$ is defined as
\begin{align}
\begin{split}
    \mathcal{I}_S[t^+_j, t^-_{j+1}] \triangleq \int_{t^+_j}^{t^-_{j+1}} & \left(\frac{1}{2} \lVert u^j_t \rVert^2 + V(t,X^j_t) dt + \right.
    \\
    & \left. \sqrt{\epsilon} (u^j_t)' d \Tilde{W}^j_t \right) + \Psi_T.
\end{split}
\end{align} 
The optimal control \eqref{eq:ustar_P0} is then equivalently
\begin{equation}
\label{eq:ustar_Pu}
u_t^* = \frac{\sqrt{\epsilon} \times \mE_{\mP^u} \left[ \Delta W_t \exp \left( -\frac{1}{\epsilon}\mathcal{S}_H(t) \right) \right] }{\Delta t \times \mE_{\mP^u} \left[ \exp \left( -\frac{1}{\epsilon}\mathcal{S}_H(t) \right) \right] }.
\end{equation}
Here $\Delta W_t$ is a discrete-time increment of the Wiener process under $\mP^0$. By Girsanov's theorem, the variable 
\begin{equation*}
    \Delta \Tilde{W}^j_t \triangleq \Delta W_t^j - \sigma_j \frac{u_t \Delta t}{\sqrt{\epsilon}}
\end{equation*}
is a Wiener process under $\mP^u$. Plugging into \eqref{eq:ustar_Pu}, we have \begin{equation}
\label{eq:optimal_control}
    u_t^{\ast} = u_t + \frac{ \sqrt{\epsilon} \times \mE_{\mP^u} \left[ \exp\left(-\frac{1}{\epsilon}\mathcal{S}_H^u(t)\right) \Delta \Tilde{W}_t \right] }{ \Delta t \times \mE_{\mP^u} \left[ \exp \left(-\frac{1}{\epsilon}\mathcal{S}_H^u(t)\right) \right]}.
\end{equation}
\eqref{eq:optimal_control} is an expectation under the proposal measure $\mP^u$, which can be obtained by an efficient sampling of the controlled SDE under $u$ with the hybrid constraints. 

\subsection{Hybrid i-LQR Proposal with Reference Extensions}
The Hybrid iLQR controller \cite{kong2021ilqr, KongHybridiLQR} solves our problem by iterative linearization. The Bellman update at $i \rm{th}$ state reads
\begin{equation}
\label{eq:belleman_update}
    \mathcal{J}^*(X^j_i) = \min_{u_i^j} \left [ l_{\Delta}(X^j_i, u^j_i) + \mathcal{J}^*(F^H_{\Delta,i}(X^j_i, u^j_i)) \right],
\end{equation}
where $l_{\Delta} \triangleq (V(X^j_i)+\frac{1}{2}\lVert u^j_i\rVert^2) \times \Delta t$ is the step running cost, and $\mathcal{J}^*$ is the value function. H-iLQR approximately solves \eqref{eq:belleman_update} by linearizing the dynamics and quadratically approximating $Q=l_{\Delta}+\mathcal{J}^*$ around a nominal sequence $\{ (\bar x_i, \bar u_i )\}_{N_T}$. Details of the backward pass in H-iLQR are presented in Appendix \ref{sec:appendix-hilqr}.

{\em Mode mismatch and reference trajectory extensions.} One issue of the feedback control law for hybrid systems is the mode mismatch \cite{rijnen2015optimal, rijnen2017control, KongHybridiLQR}, i.e., the actual and reference states may be in different modes at the proximity of the hybrid transitions. In this case, the state deviations $\delta X_t = X_t - \bar{x}_t$ have no physical meaning. Reference trajectory extensions can mitigate this issue. The stochastic smooth flow amplifies the chances of mode mismatch because of the error accumulated from the stochasticity. Consider a reset map $R_{jk}$. At time $t$, the mode mismatch has two situations. 
\begin{itemize}
    \item The \textit{early arrival}, where the true state triggers the guard function earlier than the reference, i.e., $X_t \in I_{j}, \; \bar{x}_t \in I_{k};$
    \item The \textit{late arrival}, where the reference triggers the guard function earlier than the true state, i.e., $\bar{x}_t \in I_{j}, \; X_t \in I_{k}.$
\end{itemize}

Computation details of reference extensions are presented in Appendix \ref{sec:method_mode_mismatch}.

\section{Main Algorithm}
\label{sec:method}
This section introduces the hybrid path integral with the hybrid i-LQR proposal controller control framework. 

\subsection{Main Algorithm: Hybrid Path Integral Control}
\label{sec:method_h_PI}
We present our framework, namely the Hybrid Path Integral algorithm with Hybrid iLQR Proposal (H-PI-iLQR), formalized in Algorithm \ref{alg:h-pi}. We solve the stochastic control problem for hybrid systems \eqref{eq:formulation_main}. 

At each time step, the line $3-5$ collects $K$ sampled paths and costs. The sampled costs are then used in line $7$ to update the current proposal controller computed in line $6$. Line $8$ generates the noise for the actual system. The optimal control signal is sent to the actuator under this noise in line $9$. 

Algorithm \ref{alg:rollout} specifies the stochastic hybrid rollout sub-routine. This routine differs from the smooth rollout \cite{kappen2005path} in mode mismatch and guard condition checking. At each time step, line $3-5$ checks if the current mode meets the reference mode. If not, a corresponding trajectory extension and control gains must be chosen based on the early or late arrival mismatch conditions. With the potentially modified reference and control gains, the system is pushed forward tentatively under the proposal controller in line $6-7$, with a noise generated in line $2$. We then check the guard conditions in line $5$. If the guard is triggered, the reset map is applied in line $9$ to obtain the new state and mode. We collect the running state costs in lines $14$ and the terminal cost in line $16$.

\begin{figure}
    \centering
    \includegraphics[width=0.6\linewidth]{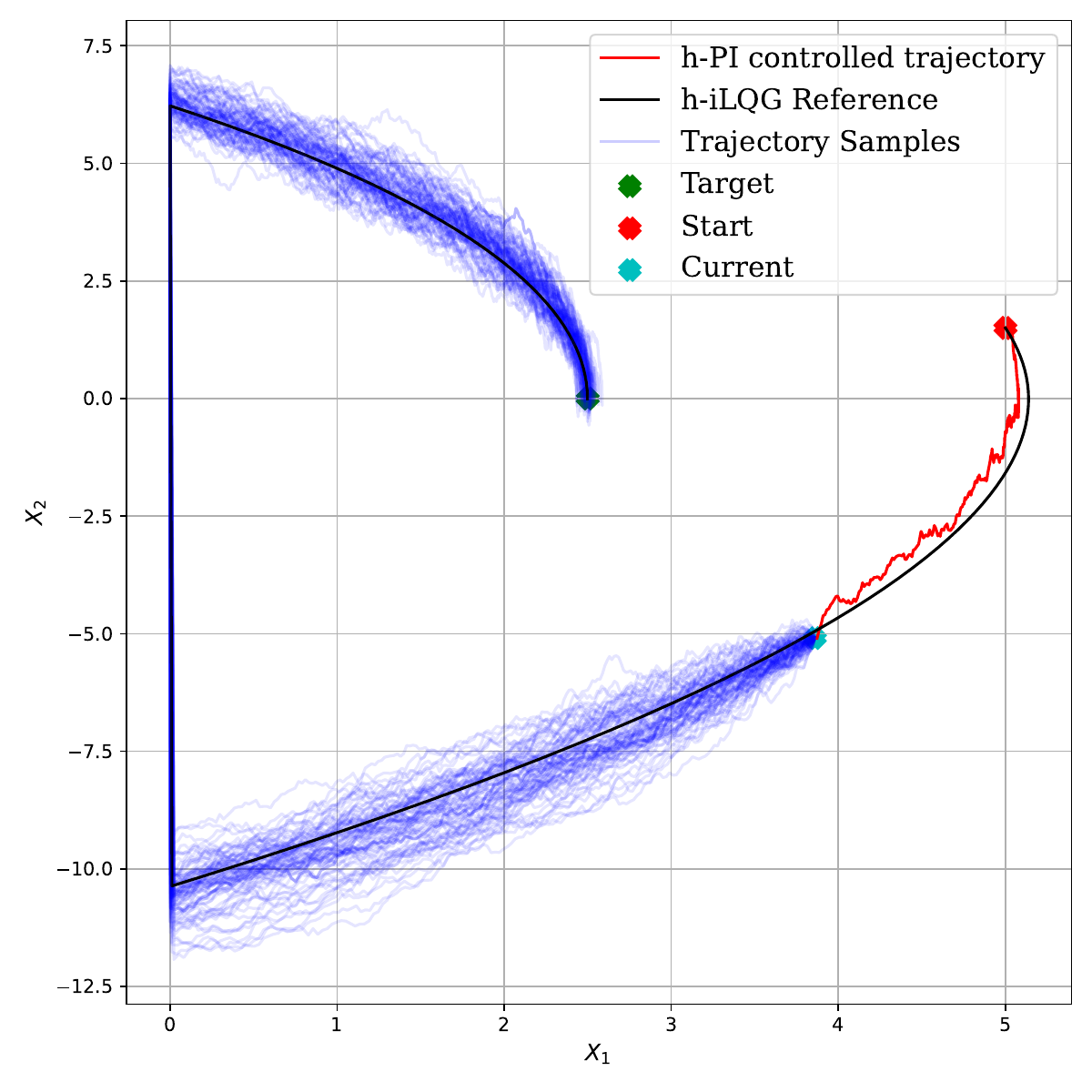}
    \caption{A visualization of the proposed H-PI-iLQR method for the bouncing ball dynamics example. }
    \label{fig:method}
\end{figure}

{
\setlength{\algomargin}{1.55em}
\begin{algorithm}[h]    
\DontPrintSemicolon
\SetAlgoLined
   
\caption{Hybrid Path Integral }
\label{alg:h-pi}
\SetKwInOut{Input}{Inputs}\SetKwInOut{Output}{Outputs}
\Input{
Initial State and Mode: $(X_0, M_0)$;  \\
Proposal Controller: \;$\Pi_{\rm Prop} = \{(\bar{x}_t,\bar{u}_t, \bar{M}_t, K_t, k_t)\}_{t=0,\dots,N_T}$;\\
Dynamics: $\Pi_{\rm Dyn} = \{F(\cdot), g(\cdot), R(\cdot)\}$\\
Other Parameters: $\Pi = \left(N_s, \beta, N_T, \Delta t, \epsilon \right )$
% \chen{$K$ is not good notation}
}
\Output{Controlled Trajectory: $\{X_t\}_{t=0, \dots, N_T}$.
}
\For{$t=0, \dots N_T - 1$}{
$\Delta W \leftarrow {\rm RandN}(N_s, N_T-t, m)$

\tcc{Hybrid Rollout (Alg. \ref{alg:rollout}).}
\For{$k=1, \dots N_s$}{
$S_k \leftarrow Rollout(X_t, M_t, \Pi_{\rm Prop}, \Pi_{\rm Dyn}, \Pi)$ 
}
$u_t \leftarrow ProposalControl(X_t, M_t, \Pi_{\rm Prop})$
\\
$u_t^{\ast}  = u_t + \frac{\sum_{k=1}^{N_s} \left ( \sqrt{\epsilon} \times \Delta W(k,0) \times \exp\left(-\frac{1}{\epsilon} S_k\right) \right ) }{ \Delta t \times \sum_{k=1}^{N_s} \left ( \exp\left(-\frac{1}{\epsilon} S_k\right) \right ) }$

\tcc{Send to Actuator.}
$d\Tilde{W}_t = \sqrt{\Delta t} \times {\rm RandN}(m,1)$  \\
$(X_t, M_t) \leftarrow HybridDyn (X_t, M_t, u_t^{\ast}, \Delta \Tilde{W}_t)$

}
\end{algorithm}
}

{
\setlength{\algomargin}{1.55em}
\begin{algorithm}[ht]    
\DontPrintSemicolon
\SetAlgoLined
   
\caption{H-iLQR Rollout with Hybrid Events}\label{alg:rollout}
\SetKwInOut{Input}{inputs}\SetKwInOut{Output}{outputs}
\Input{
Dynamics: $\{F(\cdot), g(x), R(x)\}, $ Horizon: $H$ \\ 
Proposal: $\{(\bar{x}_t,\bar{u}_t, \bar{M}_t, K_t, k_t)\}_{t=0,\dots,H}$ \\
Cost Functions: $V(t,x), \Psi_T(X_T)$ 

}
\Output{
Sample Path Cost: $S^u$.
}
\For{$t=0, \dots H - 1$}{
    $\Delta W_t = \sqrt{\Delta t}\times{\rm RandN}(m, 1)$
    \\
    % \tcc{Mode Mismatch.}
    \If{$M_t \neq \bar{M}_t$}{
    $\left(\bar{x}_t, \bar{u}_t, K_t, k_t \right) \leftarrow SelectExtension(M_t, \bar{M}_t)$ 
    }
    $u_t \leftarrow \bar u_t + K_t(X_t-\bar x_t) + k_t$    
    \\
        $\hat{X}_t  \leftarrow F(X_t, u_t, \Delta t, \Delta W_t)$      
        
        \tcc{Check guard conditions.}
        
        \If{$g(\hat{X}_t, M_t) \leq 0$}{
            $(X_t, M_t) \leftarrow R(t, \hat{X}_t)$
        }
        \Else{
            $X_t \leftarrow \hat{X}_t$;
            
        }
        $S^u \leftarrow S^u + \left (V(t, X_t) + \frac{1}{2}\lVert u_t \rVert^2\right)  \Delta t + \sqrt{\epsilon} u_t' \Delta W_t $
}
$S^u \leftarrow S^u + \Psi_T(X_T)$
\end{algorithm}
}

\subsection{Proposal Controller Quality and Sample Efficiency}
\label{sec:method_variance_reduction}
We can view the optimal controller \eqref{eq:optimal_control} as an update rule to the proposal $u_t$ as an expectation of the random variable
\[
u_t^* = u_t + \frac{\sqrt{\epsilon}}{\Delta t} \mE_{\mP^u} \left[ \alpha^u(t) \Delta \Tilde{W}_t \right] \coloneqq u_t + \Delta u_t
\]
where the weight of a trajectory $\alpha^u(t)$ is
\[
\alpha^u(t) \triangleq \frac{\exp\left(-\frac{1}{\epsilon}\mathcal{S}_H^u(t)\right) }{\mE_{\mP^u} \left[ \exp \left(-\frac{1}{\epsilon}\mathcal{S}_H^u(t)\right) \right]}
\]
The sampling efficiency of the update \eqref{eq:optimal_control} depends on the variance of $\alpha^u(t)$, which is predominantly decided by the quality of the proposal controller $u_t$. Under the optimal controller, the variance of $\alpha^{u^*}$ is zero. Indeed, we have
\begin{equation}
\label{eq:optimal_distribution_zero_variance}
    \frac{d \mP^*}{d \mP^u} = \frac{d \mP^*}{d \mP^0} \frac{d \mP^0}{d \mP^u} \propto \exp\left(-\frac{1}{\epsilon} S^u(t)\right), 
\end{equation}
and when $u_t = u^*_t$, the right-hand side of \eqref{eq:optimal_distribution_zero_variance} is $1$, and thus $\alpha^{u^*}(t)$ is an uniform distribution with variance $0$. The proposal controller quality can also be measured by the effective sample portion of $\alpha^u(t)$ using the following quantity \cite{kappen2005path} 
\[
\lambda^u \triangleq 1 / \mE\left[ \left(\alpha^{u}\right)^2 \right].
\]

\section{experiments}
\label{sec:experiments}
This section presents our experiment results to validate the proposed methods. We choose bouncing ball dynamics with linear smooth flows and elastic impact as a canonical experiment setting. Next, we evaluate the proposed method on a classical hybrid system representing simple running and walking with smooth nonlinear flows. Finally, we do extensive ablation studies on the proposed H-PI method. 

\subsection{Bouncing Ball with Elastic Impact}
\label{sec:experiments_bouncingball}
We begin with a 1D bouncing ball dynamics under elastic impact. The system has $2$ modes, $\mathcal{I} = \{I_1, I_2\}$ where $I_1 \triangleq \{ z | \dot z < 0 \}, \; I_2 \triangleq \{ z | \dot z \geq 0 \}.$ The state in the $2$ modes has the same physical meaning and consists of the vertical position and velocity $X^0=X^1\triangleq[z, \dot z]'$. The control input is a vertical force. 
The system has the same flow 
\begin{equation}
    \!\!F_j \triangleq dX^j_t = \left(\begin{bmatrix}
        0 & 1
        \\
        0 & 0
    \end{bmatrix}X^j_t + \begin{bmatrix}
        0 \\ \frac{u^j_t-mg}{m}
    \end{bmatrix} \right) dt + \sqrt{\epsilon}\begin{bmatrix}
        0 \\ \frac{1}{m}
    \end{bmatrix} dW^j_t.
\end{equation}
The guard functions are defined as $g_{12}(z,\dot z) \triangleq z$, and $g_{21} \triangleq \dot z$. The reset map $R_{21}$ is the identity map, and 
\begin{equation}
    \begin{bmatrix}
     z(t^+) 
     \\
     \dot z(t^+)
    \end{bmatrix}
    =
    R_{12}(z(t^-), \dot z(t^-)) \coloneqq 
    \begin{bmatrix}
    1 & 0 
    \\
    0 & -e_2
    \end{bmatrix}
    \begin{bmatrix}
     z(t^-) 
     \\
     \dot z(t^-)
    \end{bmatrix}
    \label{eq:reset_map_bouncing}
\end{equation}
where $e_2$ represents the elastic loss in position and velocity at the contact. The start and goal states are $x_{S} = [5, 1.5]'; \; x_{G} = [2.5, 0]'$, both in mode $I_2$. In \eqref{eq:formulation_main}, we use a quadratic terminal loss $\Psi_T(X_T) \coloneqq \frac{1}{2} \lVert X_T \rVert^2_{Q_T}$ where $Q_T=\begin{bmatrix}
    200.0 & 0.0
    \\
    0.0 & 20.0
\end{bmatrix}.$
All experiments use $\Delta t=0.0025, \epsilon=10.0$, and we use $5000$ trajectory samples for a path integral control update.

% {\em a) Improvement in the expected costs.}

We evaluate the expected cost $\mathcal{J}_H$ in \eqref{eq:formulation_main} using a Monte Carlo estimation. Specifically, we conduct $100$ experiments under independent randomness and compute the empirical mean of the expected path costs. We use the same randomnesses to compare the expected path costs for the H-PI controller and the H-iLQR proposal controller. 

\begin{figure}[t]
\centering
\begin{subfigure}{0.241\textwidth}
    \includegraphics[width=\linewidth]{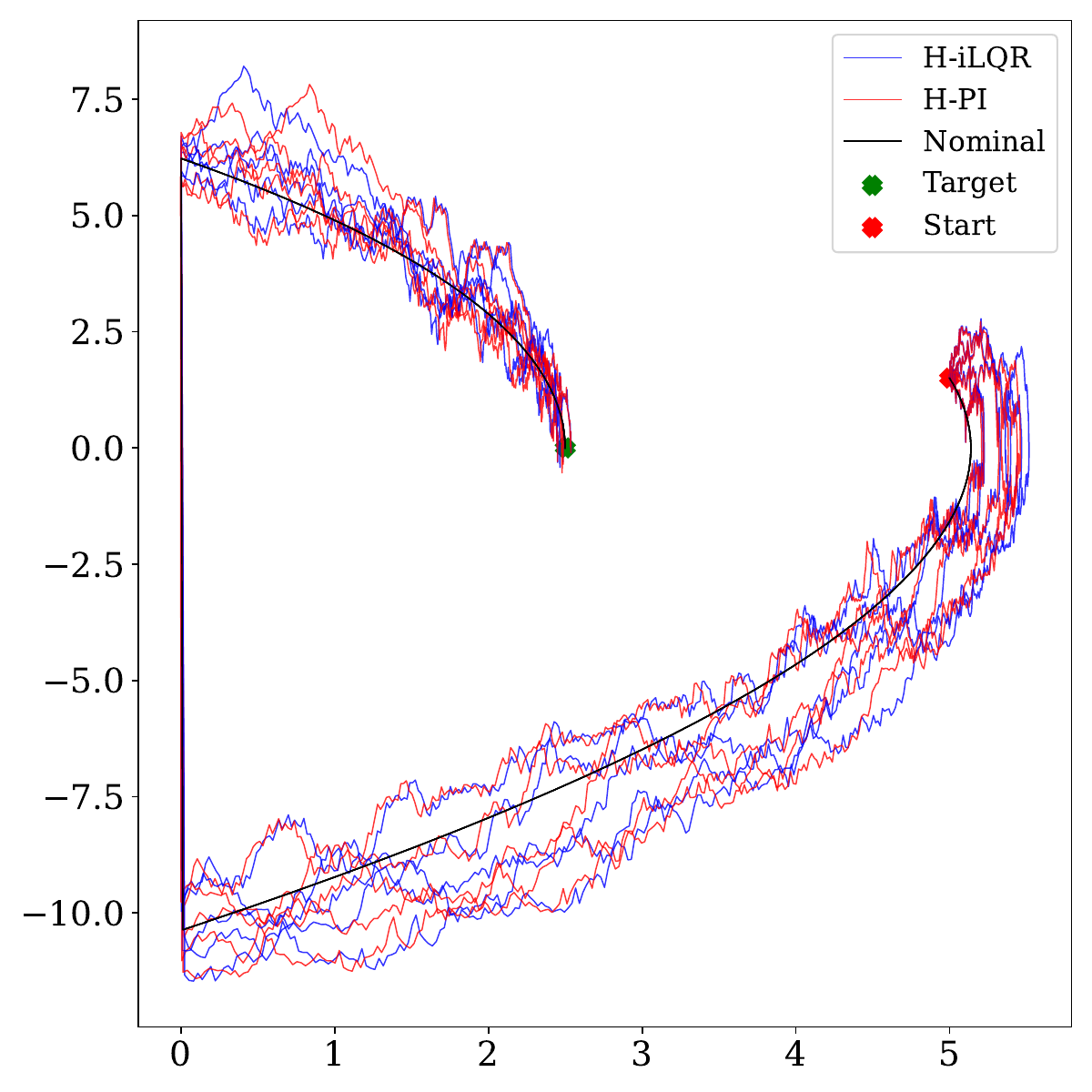}
    \caption{Trajectories of the best $5\%$ H-iLQR proposal cost.}
    \label{fig:bouncing_best_10}
\end{subfigure}
\hfill
\begin{subfigure}{0.241\textwidth}
    \includegraphics[width=\linewidth]{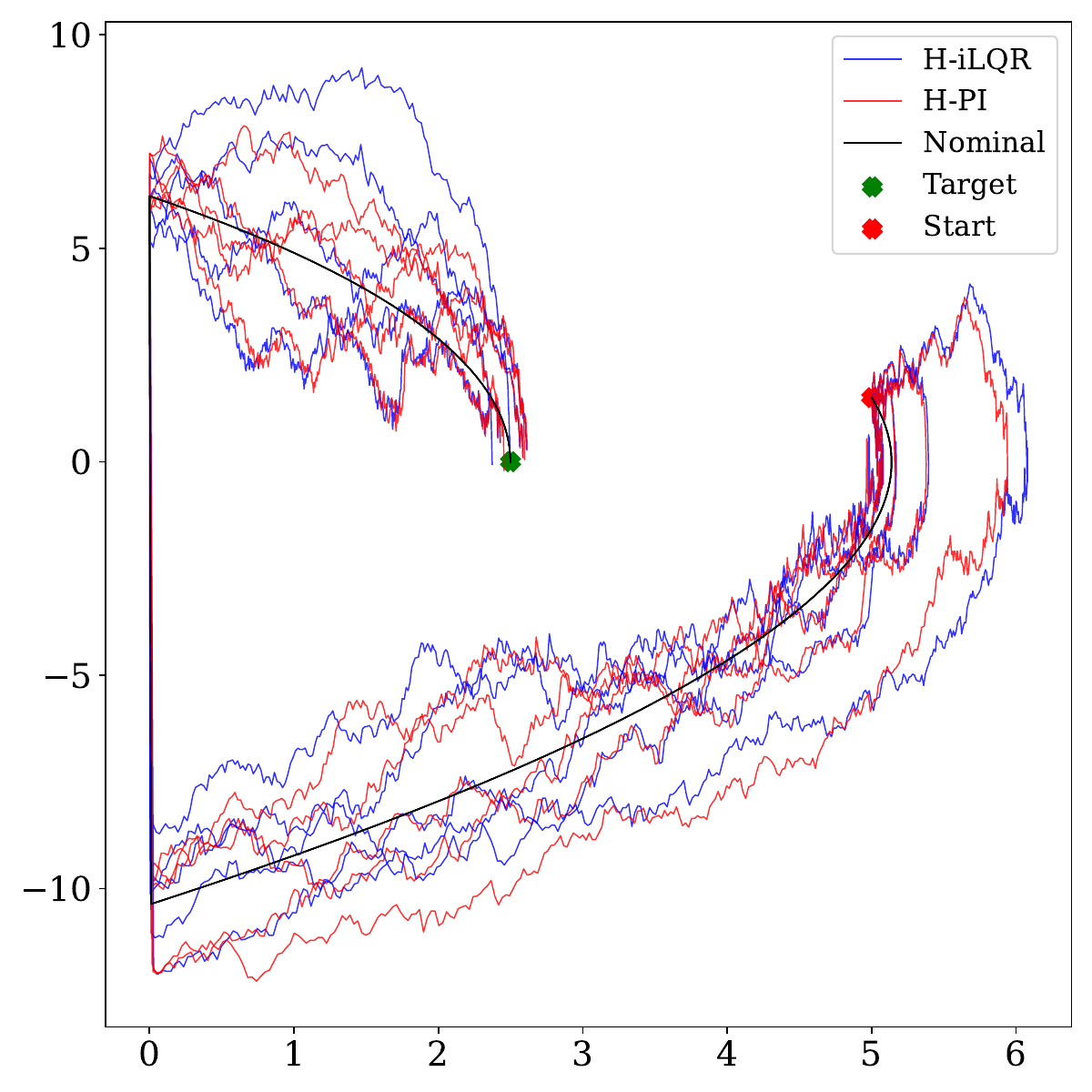}
    \caption{Trajectories of the tail $5\%$ H-iLQR proposal cost.}
    \label{fig:bouncing_tail_10}
\end{subfigure}
\caption{Controlled trajectories under H-PI and H-iLQR for the best and tail $5 \%$ costs attained by H-iLQR. Near the jumping events, larger deviations appear under H-iLQR, inducing larger control energy costs.}
\label{fig:bouncing}
\end{figure}

We plot the controlled trajectories under both H-PI and H-iLQR, for the experiments with the best and worst $5\%$ costs H-iLQR has achieved in Fig. \ref{fig:bouncing_best_10}. In the best cases, the trajectories have similar performances. In contrast, in the worst cases, the H-PI drags back the H-iLQR controlled trajectories far from the nominal, decreasing the control energy costs. 

\begin{table*}[t]
\centering
\begin{tabular}{|c|c|c|c|c|}
\hline
                  &  & H-iLQR  & H-PathIntegral & Improved $(\%)$  \\ \hline
\multirow{2}{*}{\centering Expected Cost} & Bouncing Ball & $60.11$ & $57.86$ & $3.74$   \\ \cline{2-5} 
                  & SLIP & $0.2459$  & $0.2164$ & \textbf{12.00} \\ \hline
\multirow{2}{*}{\centering H-iLQR $10\%$ tail} & Bouncing Ball &   $141.65$  &   $115.40$   &   $\textbf{11.46}$   \\ \cline{2-5} 
                  & SLIP & $0.9451$ & $0.2616$ & $\textbf{63.29}$ \\ \hline
\end{tabular}
\caption{Expected cost improvement and the statistics on the experiments where H-iLQR has the highest $10\%$ costs. }
\label{tab:bouncing}
\end{table*}

\begin{table*}[t]
    \centering
    \begin{tabular}{|c|c|c|c|c|}
    \hline
    & Confidence level & $0.7$ & $0.8$ & $0.9$ 
    \\
    \hline
    \multirow{2}{*}{\centering Improvement $(\%)$} & Bouncing Ball & $15.05$ & $21.34$ & $32.44$
    \\ \cline{2-5}
             &  SLIP & $37.86$ & $47.48$ & $59.32$
    \\ \hline
    \end{tabular}
    \caption{Conditional Value at Risk (CVaR) statistics for the cost improvement in Tab. \ref{tab:bouncing}, conditioned on confidence levels $70\%$, $80\%$, and $90\%$. } 
\label{tab:CVaR_bouncing}
\end{table*}

\begin{figure*}[t]
    \centering
    \includegraphics[width=0.195\linewidth]{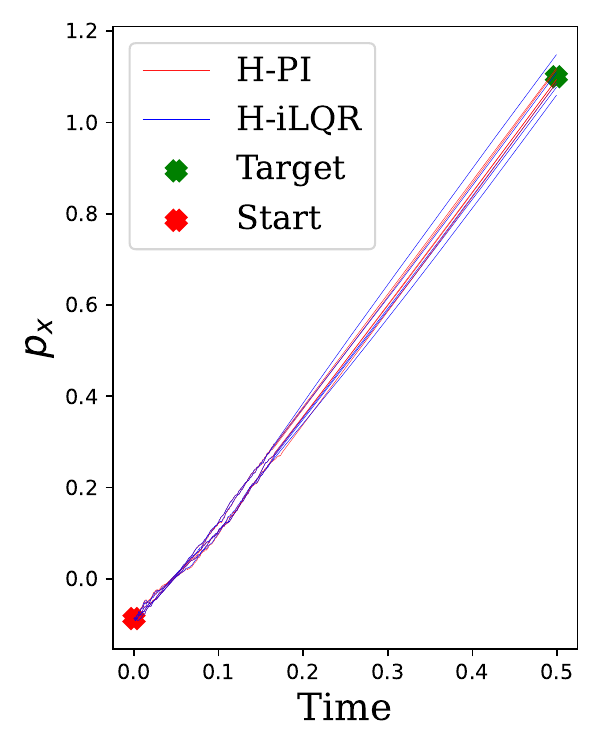}
    \includegraphics[width=0.195\linewidth]{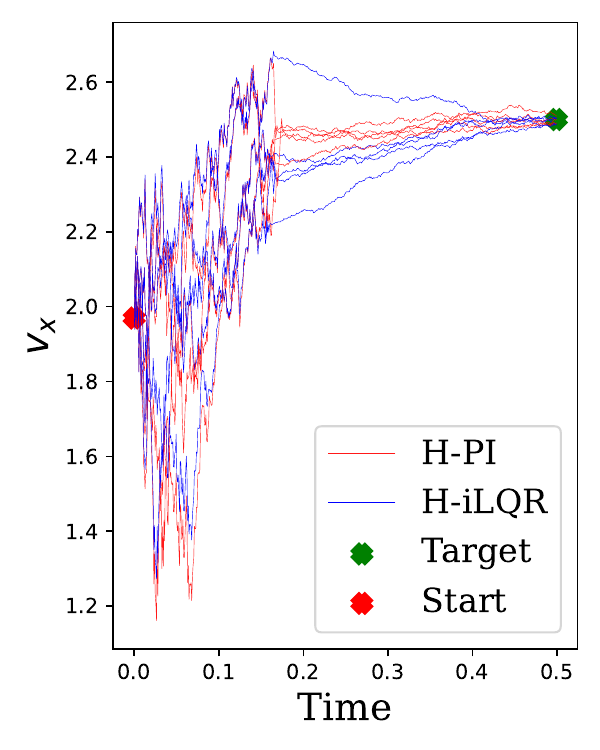}
    \includegraphics[width=0.195\linewidth]{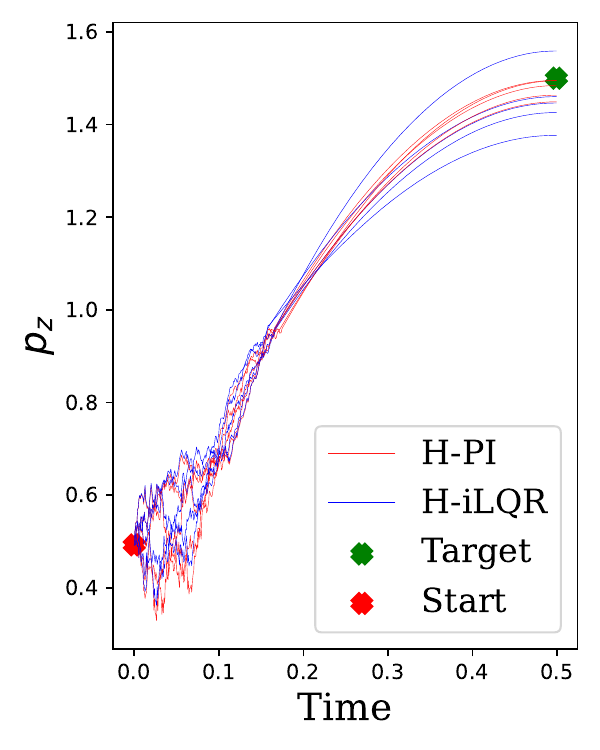}
    \includegraphics[width=0.195\linewidth]{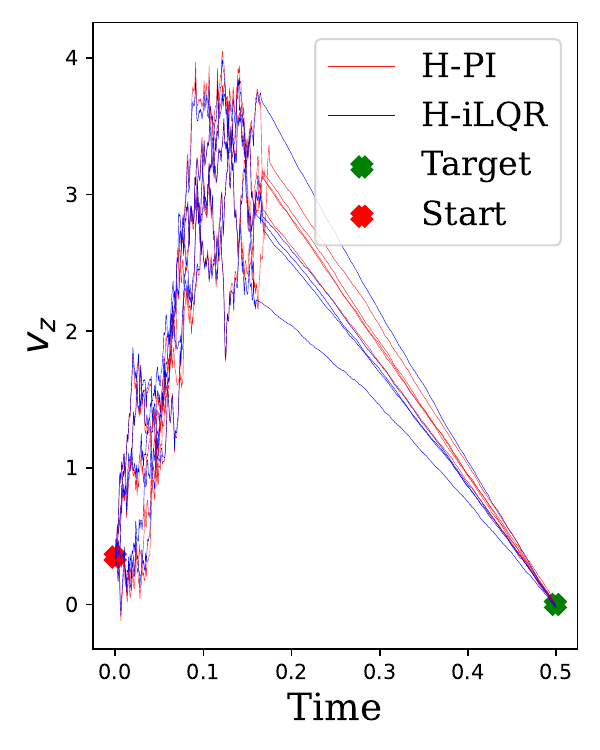}
    \includegraphics[width=0.195\linewidth]{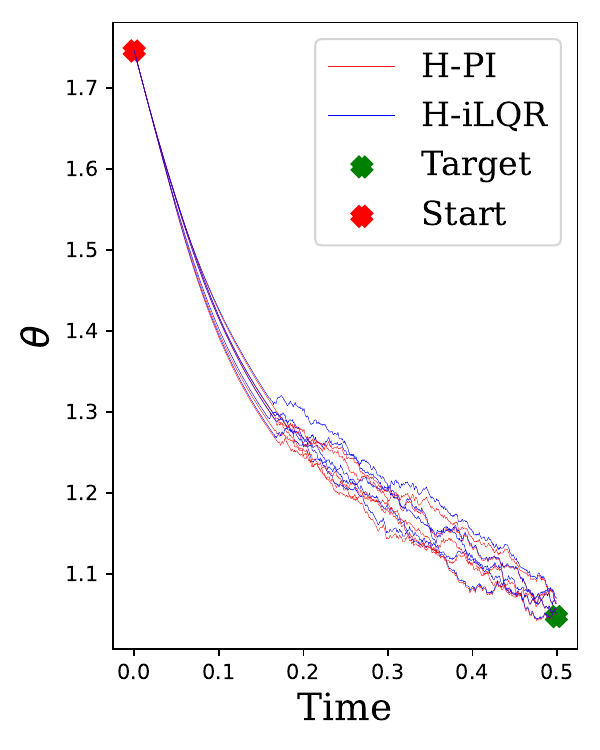}
    \caption{Controlled trajectories of the SLIP jumping experiment, for the tail $5\%$ cost attained by H-iLQR.}
\label{fig:slip_best_trj}

\caption{Trajectories under the same randomness, controlled by H-PI and H-iLQR for the tail $5 \%$ costs attained by H-iLQR in the $100$ experiments. }
\label{fig:slip_result}
\end{figure*}

\subsection{Spring-Loaded Inverted Pendulum (SLIP) Model}
\label{sec:experiments_slip}
We test our algorithm on a classical spring-loaded inverted pendulum (SLIP) nonlinear dynamics. The SLIP model consists of a body modeled by a point mass $m$ and a leg modeled by a massless spring with coefficient $k$ and natural length $r_0$. We use $(p_x, p_z)$ to represent the horizontal and vertical positions of the body, and $(v_x, v_z)$ are their velocities, respectively. $(\theta, \dot \theta)$ are the angle and angular velocity between the leg and the horizontal ground, and $(r, \dot r)$ are the leg length and its changing rate. The variables $[p_x, v_x, p_z, v_z, \theta, \dot \theta, r, \dot r]$ fully describe the system's states.  

The SLIP system has two modes, denoted as $\mathcal{I} = \{I_1, I_2\}$ where $I_1 \triangleq \{ X^1_t | p_z - r_0 \sin{\theta} \geq 0 \}, \; I_2 \triangleq \{ X^2_t | p_z - r_0 \sin{\theta} < 0 \}$. $I_1$ is also known as the \textit{flight} mode, and mode $I_2$ is defined as the \textit{stance} mode. We use a polar coordinate system for the stance mode, leading to different state spaces in the two modes. In the flight mode, the state space is $X^1_t = [p_x, v_x, p_z, v_z, \theta]$, and the smooth flow is defined as
\begin{equation}
    \label{eq:dyn_SLIP_F1}
    d X^1_t = 
    \left(\begin{bmatrix}
        \dot v_x\\
         0\\
        \dot v_z\\
         -g\\
        0
    \end{bmatrix}
    +
    \begin{bmatrix}
        0\\
         u^1_{1,t}\\
        0\\
         u^1_{2,t}\\
        u^1_{3,t}
    \end{bmatrix}
    \right) dt + \sqrt{\epsilon}\begin{bmatrix}
        0 & 0 & 0\\
        1 & 0 & 0\\
        0 & 0 & 0\\
        0 & 1 & 0\\
        0 & 0 & 1
    \end{bmatrix}
    dW^1_t
\end{equation}
where $g$ is the gravity and $u^1_t \in \mR^3$ is the control input in mode $1$. We assume the system controls its body's and leg's velocities in the flight mode.

In the stance mode, the state is represented by $X^2_t = [\theta, \dot \theta, r, \dot r]$, and the stance smooth flow is
\begin{equation}
    \label{eq:dyn_SLIP_stance}
    \begin{split}
        \!\!
    dX^2_t
    =
    &\left(\begin{bmatrix}
        \dot \theta\\
        \frac{-2\dot \theta\times \dot r - g\times \cos\theta}{r}\\
        \dot r\\
        \frac{k(r0-r)}{m} - g\sin\theta + \dot \theta^2 r
    \end{bmatrix}
    +
    \begin{bmatrix}
        0 \\
        0 \\
        \frac{m}{r^2}u^2_{1,t} \\
        \frac{k}{m}u^2_{2,t} 
    \end{bmatrix}
    \right) d t
    +
    \\
    &\sqrt{\epsilon}
    \begin{bmatrix}
        0 & 0 \\
        0 & 0 \\
        \frac{m}{r^2} & 0  \\
        0 & \frac{k}{m} 
    \end{bmatrix}
    \begin{bmatrix}
        dW^2_{1,t}
        \\
        dW^2_{2,t}
    \end{bmatrix}.
    \end{split}
\end{equation}

The reset maps between the two modes are defined as 
\begin{equation}
\label{eq:reset_slip_12}
\!\!X^2_{t^+} = R_{12}(X^1_{t^-}) = 
\begin{bmatrix}
    \theta^- \\
    (p_x^- v_z^- - p_z^- v_x^-) / r_0^2 \\
    r_0\\
    -v_x^-\cos \theta^- + v_z^-\sin\theta^-
\end{bmatrix}
\end{equation}
and 
\begin{equation}
\label{eq:reset_slip_21}
\!\!X^1_{t^+} = R_{21}(X^2_{t^-}) = 
\begin{bmatrix}
    p_{x,T}^- + r_0\cos\theta^- \\
    \dot{r}^-\cos\theta^- - r^-\dot{\theta}^-\sin\theta^- \\
    r_0\sin\theta^-\\
    r_0\dot\theta\cos\theta^- + \dot{r}^-\sin\theta^-\\
    \theta^-
\end{bmatrix},
\end{equation}
where $p_{x,T}$ is the tole $x$-potision and is assumed to be unchanged during the stance mode.

\begin{figure}
    \centering
    \begin{subfigure}[b]{0.241\textwidth}
        \includegraphics[width=\linewidth]{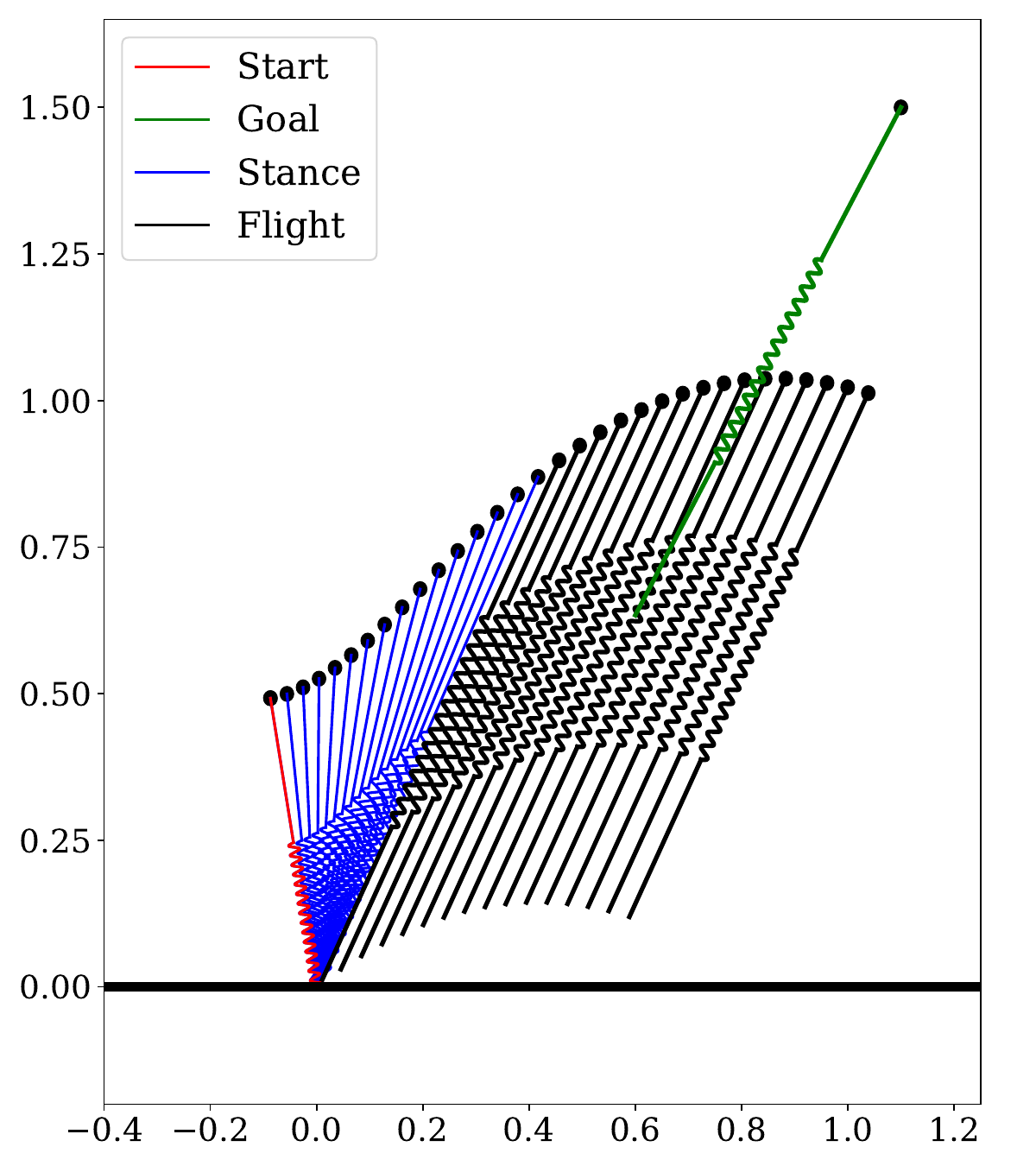}
    \caption{Initial trajectory rollout.}
    \end{subfigure}
    \hfill
    \begin{subfigure}[b]{0.241\textwidth}
        \includegraphics[width=\linewidth]{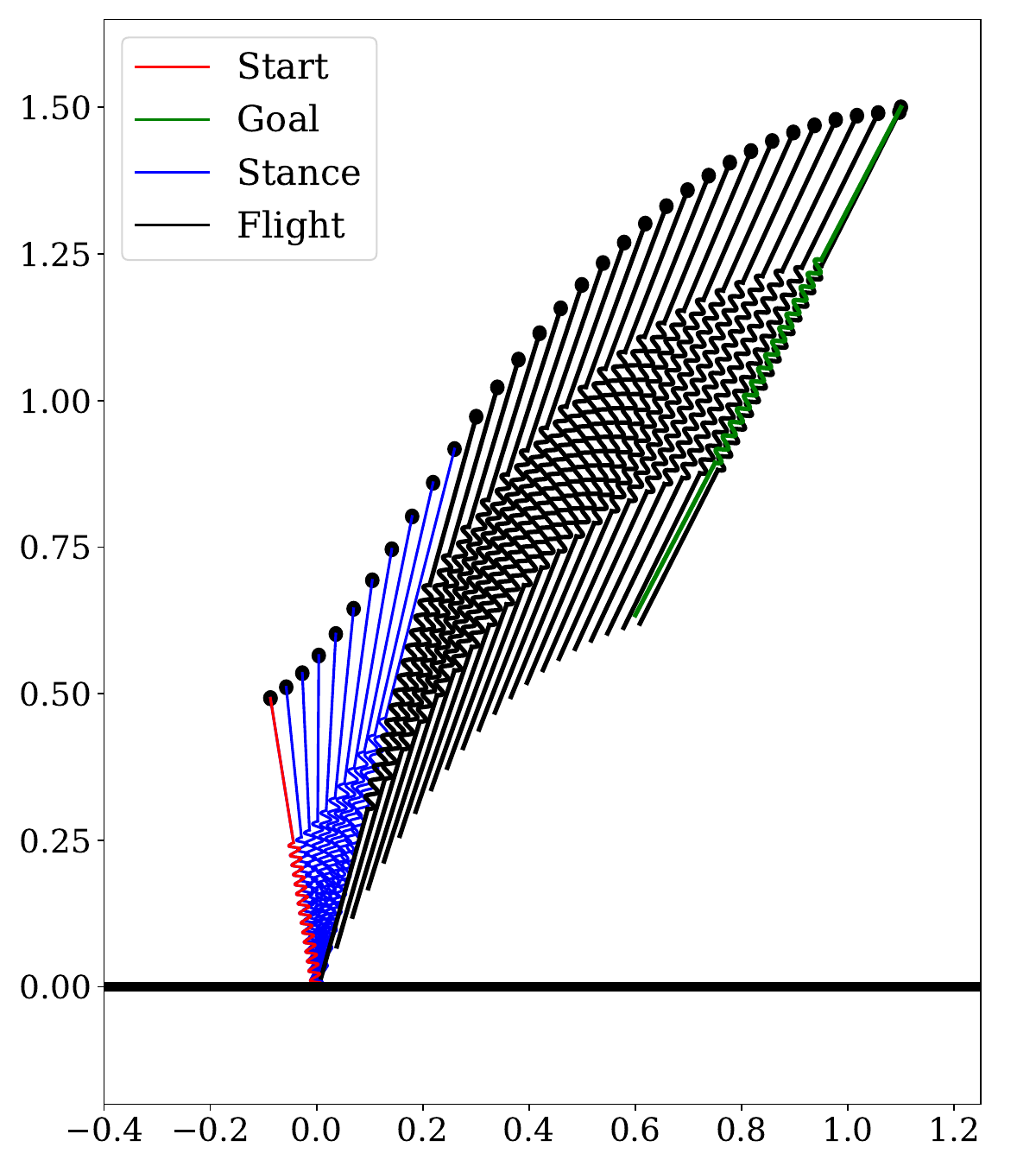}
    \caption{H-iLQR reference trajectory.}
    \end{subfigure}
    \caption{Experiment settings for the SLIP jumping task, we plot the rollout under initial guess (left) and the optimized H-iLQR reference (right).}
    \label{fig:slip_jump_setting}
\end{figure}

{\em SLIP Jumping Experiment.}
We do an experiment simulating a one-step jump using the SLIP model. In this experiment, we start from a stance position, squeezing the spring and letting go of the system to let it jump to a designated pose in flight mode. The target pose is higher than the pose reached by the system under no control. We let $X^2_0 = [1.74533, -4.0, 0.5r_0, 0.0]$, and let $X^1_T = [1.1, 2.5, 1.5, 0.0, \pi/3]$. In \eqref{eq:formulation_main}, we use a quadratic terminal loss $\Psi_T(X_T) \coloneqq \frac{1}{2} \lVert X_T \rVert^2_{Q_T}$ where $Q_T = 60 \times I_5$.
All experiments use $\Delta t=0.0008, \epsilon=0.005$, and we use $5000$ trajectory samples for a path integral control update. The experiment setting is shown in Fig. \ref{fig:slip_jump_setting}.

\begin{figure}
    \centering
    \begin{subfigure}[b]{0.241\textwidth}
        \includegraphics[width=\linewidth]{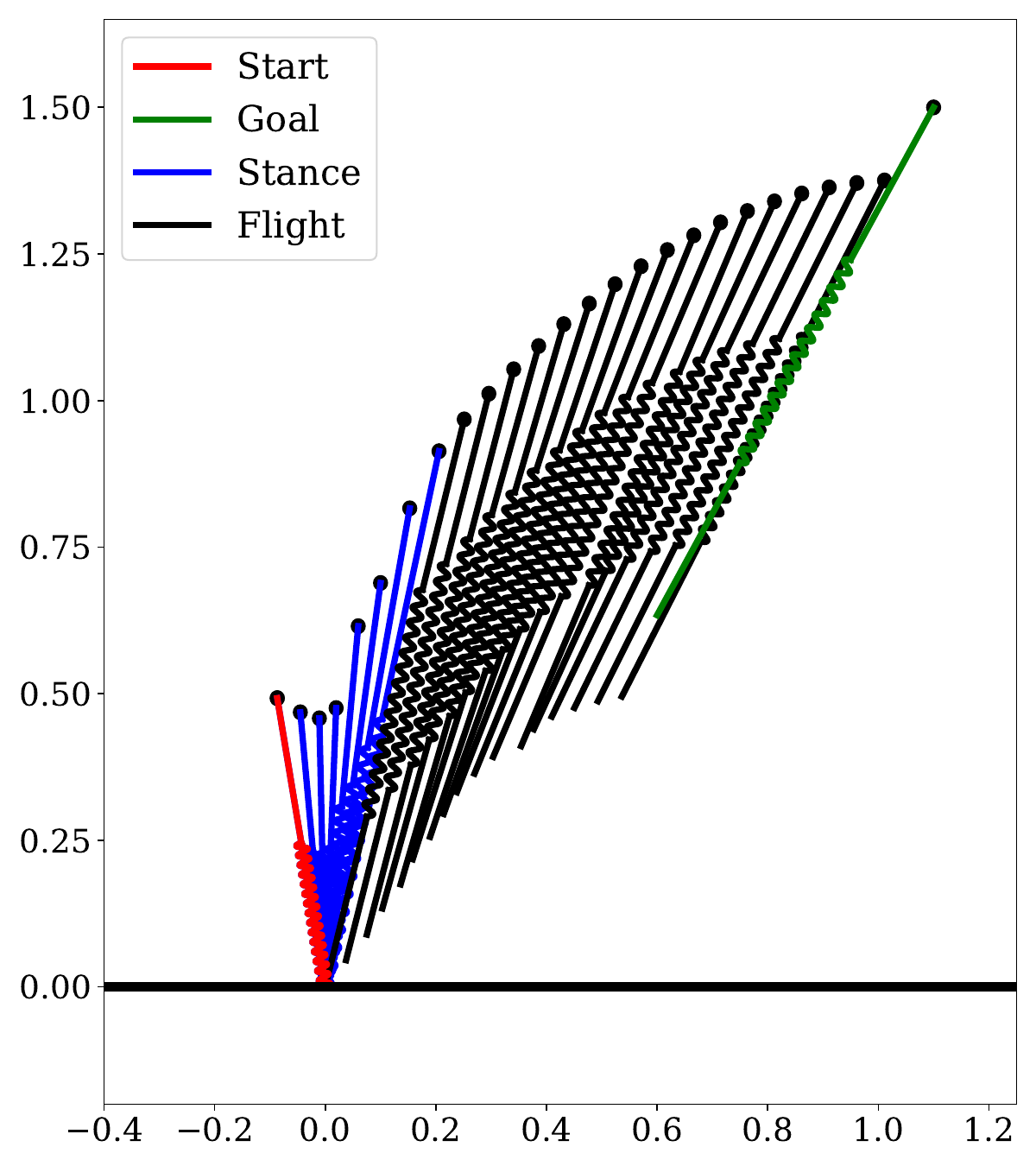}
    \caption{H-iLQR controlled trajectory.}
    \end{subfigure}
    \hfill
    \begin{subfigure}[b]{0.241\textwidth}
        \includegraphics[width=\linewidth]{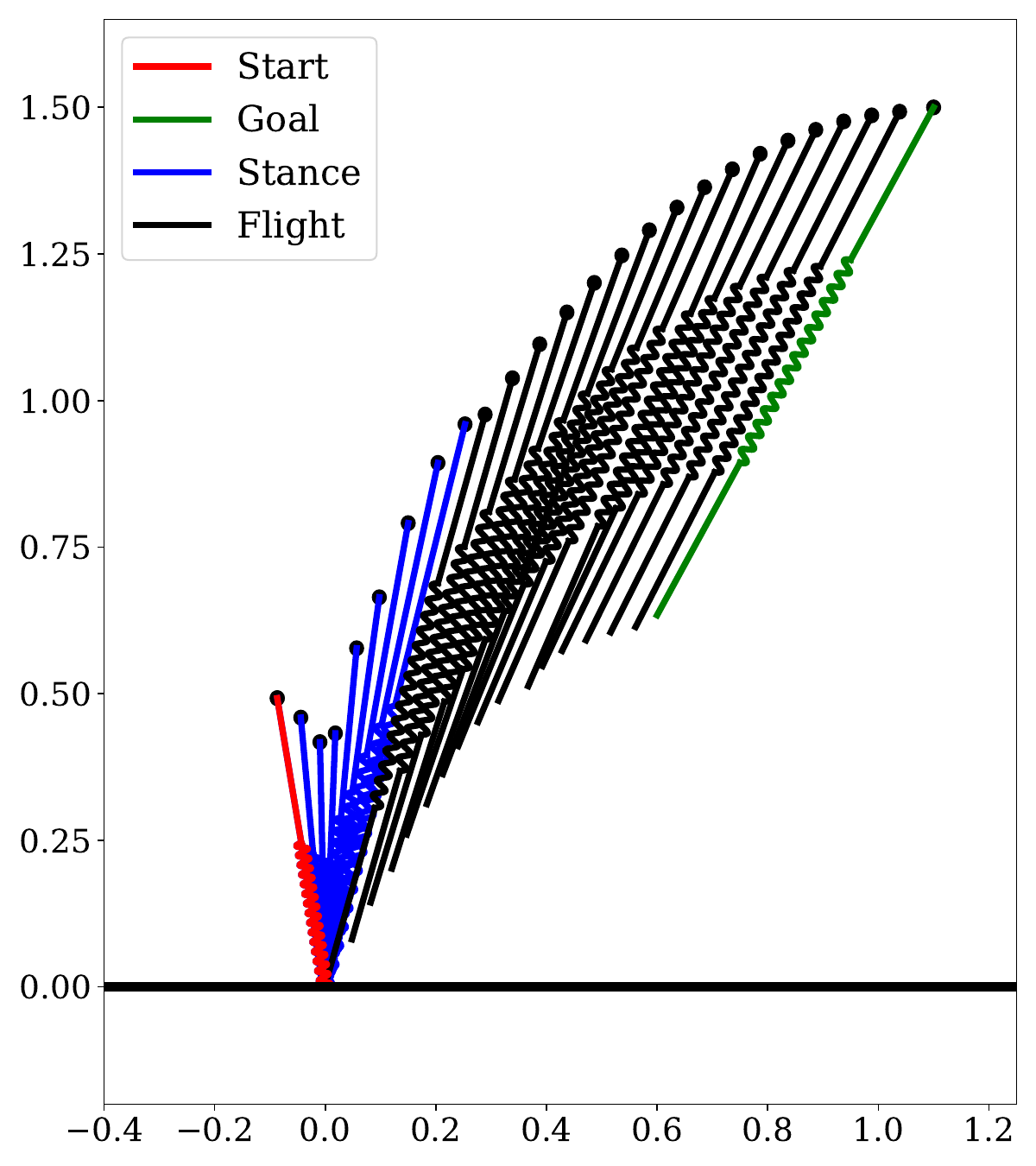}
    \caption{H-PI controlled trajectory.}
    \end{subfigure}
    \caption{Improvement of the H-PI controller in reaching the target states under the same disturbances.}
    \label{fig:animate_trajectories}
\end{figure}

The resulting controlled trajectories are shown in Fig. \ref{fig:slip_result}. As in the bouncing ball experiment, we plot the trajectories for the experiments where the H-iLQR controlled achieved its best and tail $5 \%$ costs. In Fig. \ref{fig:slip_result}, we observe sharp oscillations in the velocities $v_x$ and $v_z$ in the stance mode. In this mode \eqref{eq:dyn_SLIP_stance}, the noise $dW^2_t$ enters the SLIP system via the $r$ and $\dot r$ channels. The noises are amplified by the nonlinear coupling between $\dot{r}$ and $(v_x, v_z)$ through the length of the spring and its angular velocity. The improvements in the expected costs are recorded in Tab. \ref{tab:bouncing}. Fig. \ref{fig:animate_trajectories} shows one example experiment where the H-PI improves performance in reaching the goal states compared with the H-iLQR proposal. 

\subsection{Expected Cost Reduction Result Analysis.} The improvement in the expected costs for both the bouncing ball and the SLIP tasks are shown in Tab. \ref{tab:bouncing}. The improvement in the expected cost of the $100$ experiments is $3.74\%$ for the bouncing ball task and $12.00\%$ for the SLIP task. The difference is that, the nonlinearity in the bouncing ball dynamics comes solely from the hybrid transition (Although the reset map \eqref{eq:reset_map_bouncing} is a linear mapping, it results in discontinuous state changes), while the SLIP model has both nonlinear smooth flow and nonlinear reset maps.  

The H-PI is a policy improvement method for any given proposal control (See also Appendix \ref{sec:experiments_ablationstudy}-(b) to see an example improvement over a zero-control proposal). For the H-iLQR proposal, the expected cost improvement is largely from improving a set of high-cost trajectories (worst-cases) under uncertainties. 

To see the improvement in the high-cost regions, we compute the expected cost improvement corresponding to the trajectories with the highest $10\%$ costs under H-iLQR among the $100$ experiments for both tasks. The results are recorded in Tab. \ref{tab:bouncing}. In the $10\%$ high-cost regions, the H-PI gained $11\%$ improvement for the bouncing ball task and $63.29\%$ improvement for the SLIP task. 

We also compute the Conditional Value at Risk (CVaR) statistics of the expected cost improvement in Tab. \ref{tab:CVaR_bouncing}. For the bouncing ball task, we observe $15.05\%$, $21.34\%$, and $32.44\%$ improvements in expected costs for the $70\%$, $80\%$, and $90\%$ tail distribution, respectively. For the SLIP jumping task, the CVaRs are $37.86\%$, $47.48\%$, and $59.32\%$, for the same confidence levels.

\begin{figure*}[t]
\centering
    \begin{subfigure}[b]{0.245\textwidth}
    \includegraphics[width=\linewidth]{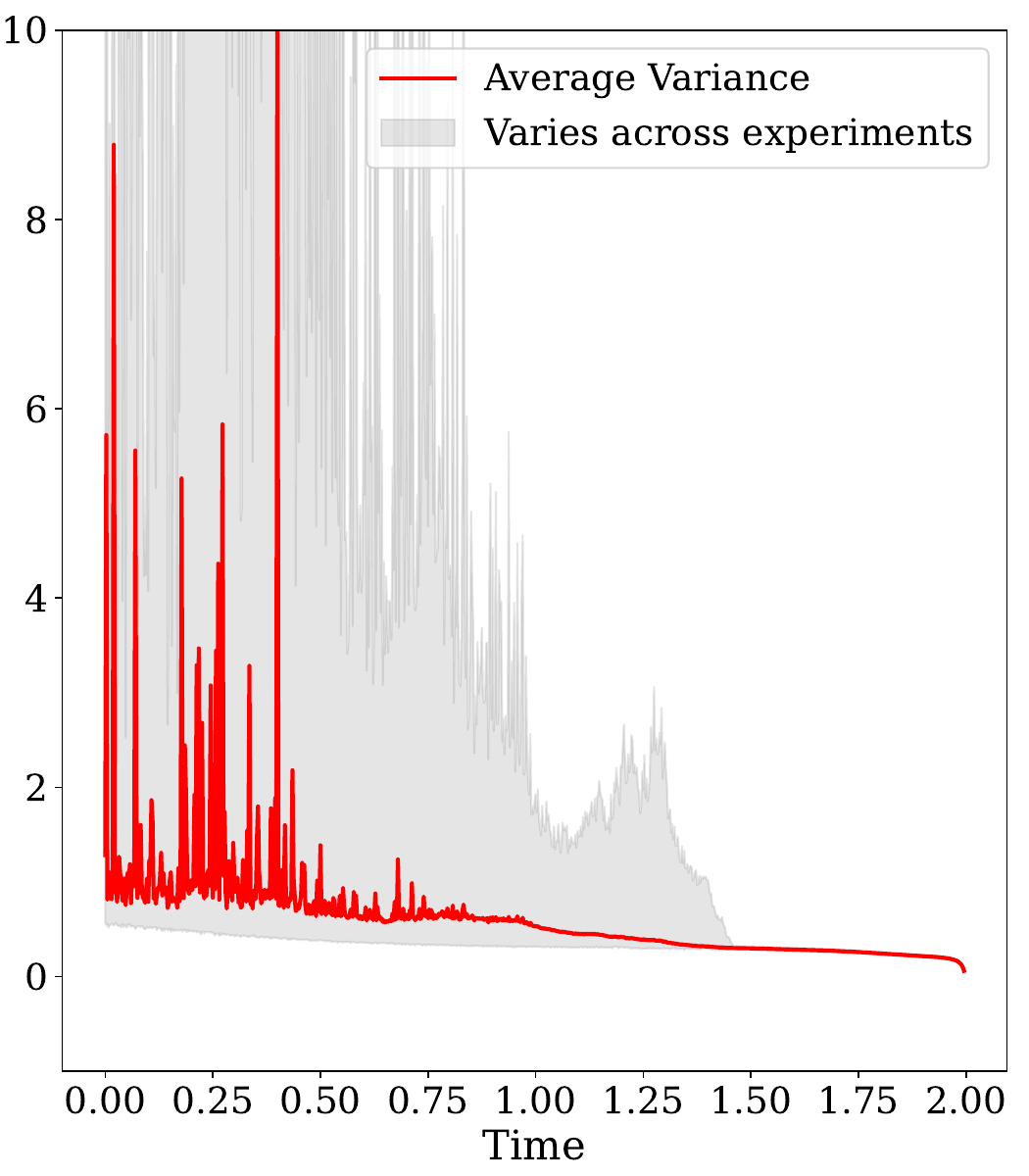}
    \caption{Variance, bouncing ball.}
    \label{fig:bouncing_var}
    \end{subfigure}
    \begin{subfigure}[b]{0.245\textwidth}
    \includegraphics[width=\linewidth]{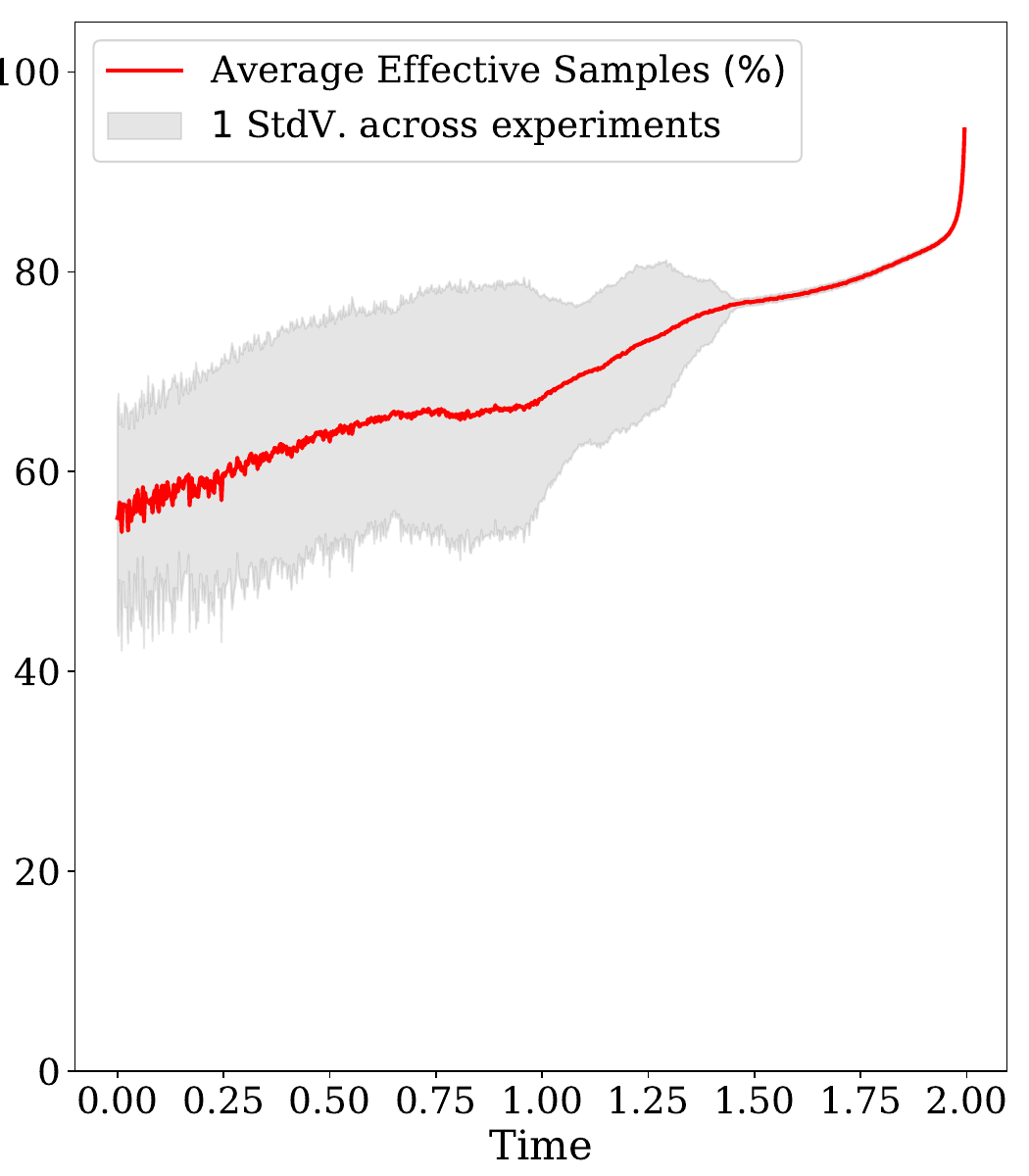}
    \caption{Effective samples, ball.}
    \label{fig:bouncing_lbd}
    \end{subfigure}
    \begin{subfigure}[b]{0.245\textwidth}
    \includegraphics[width=\linewidth]{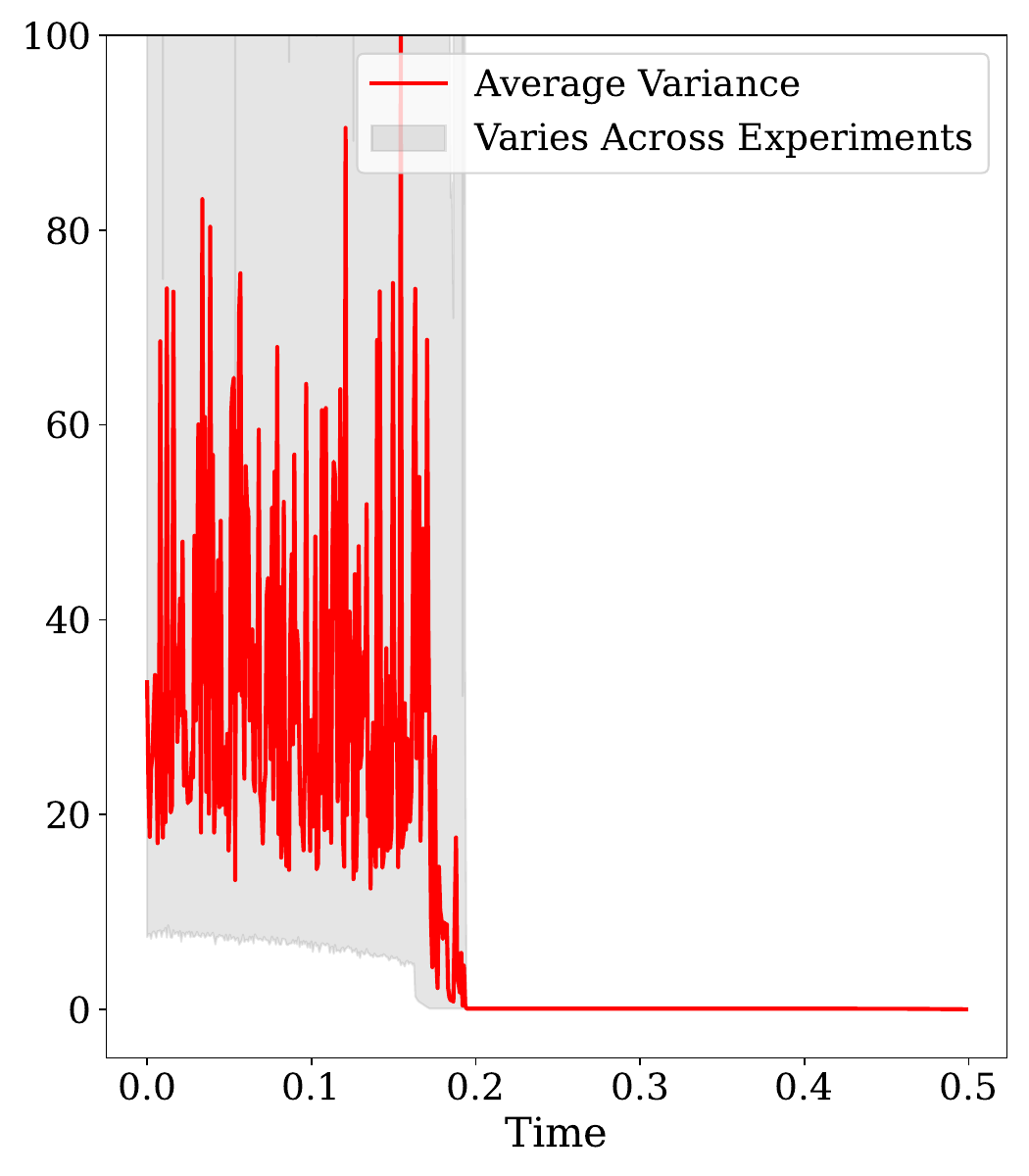}
    \caption{Variance, SLIP.}
    \label{fig:slip_var}
    \end{subfigure}
    \begin{subfigure}[b]{0.245\textwidth}
    \includegraphics[width=\linewidth]{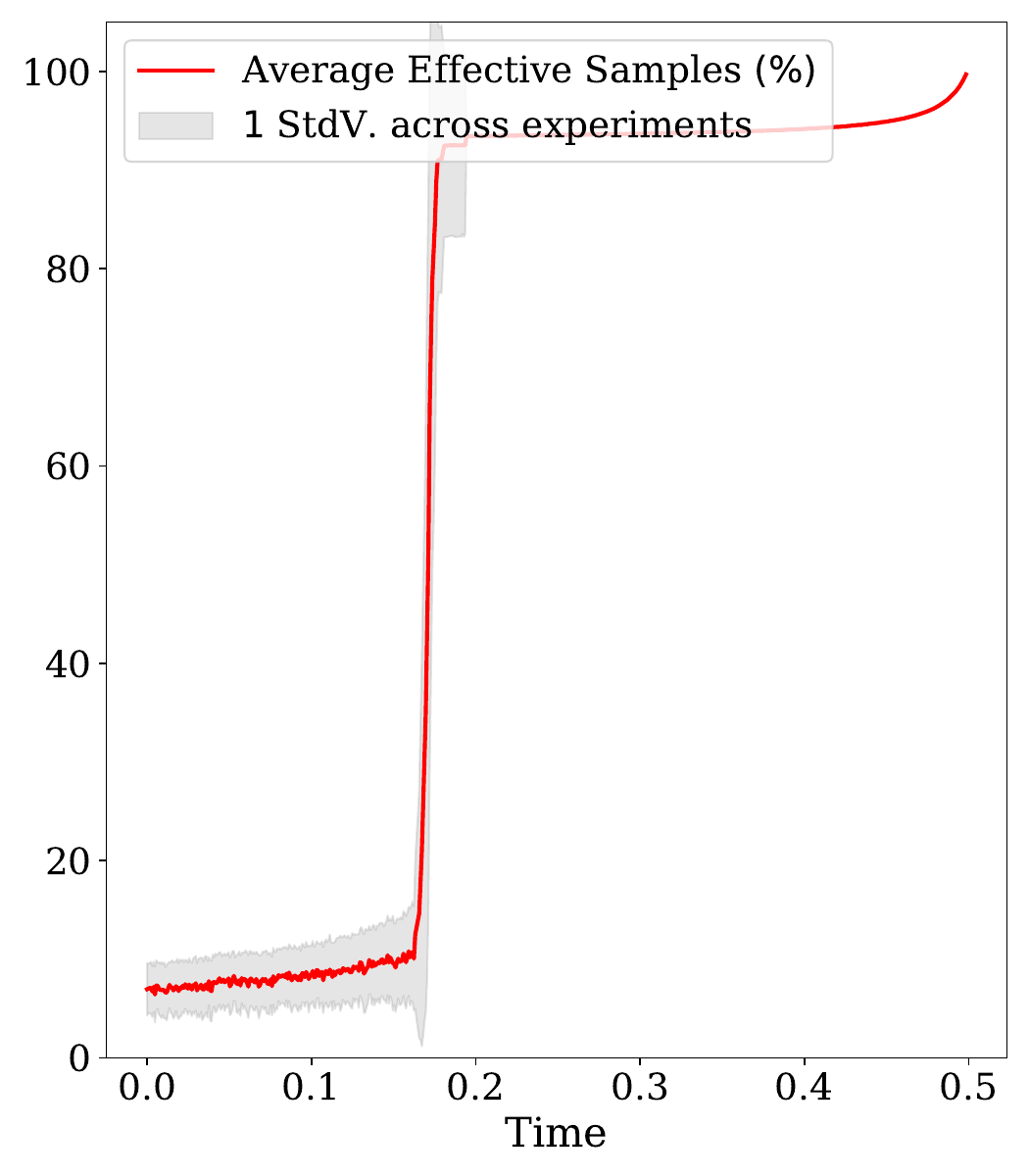}
    \caption{Effective samples, SLIP.}
    \label{fig:slip_lbd}
    \end{subfigure}
    \caption{Sample variance and effective sample portions for the bouncing ball and the SLIP jumping tasks. Both indicators experience sharp changes before and after the jump events, showing the impact of the jump dynamics on the stabilizing capability of the feedback controller. The linearization errors and the mode inconsistencies reduced the quality of H-iLQR to around $10 \%$ measured by the effective sample portion $\lambda$ for the nonlinear SLIP system.}
    \label{fig:bouncing_slip_var_lbd}
\end{figure*}
\begin{table*}[th]
\centering
\begin{tabular}{|c|c|c|c|c|c|}
\hline
&  & $[0, T]$ & $[0,t^-]$ & $[t^-, T]$ & \textbf{Jump Influence}  \\ \hline
\multirow{2}{*}{\centering Bouncing Ball}   & Avg. Var$(\alpha)$ & $0.64$ & $0.78$ & $0.25$ &  $67.95 \%$ \\ \cline{2-6} 
& Avg. $\lambda (\%)$ & $69.59$ & $65.83$ & $79.85$ & $21.2 \%$ \\ \hline
\multirow{2}{*}{\centering SLIP}   & Avg. Var$(\alpha)$ & $14.20$ & $34.03$ & $3.55$ &  $89.57 \%$ \\ \cline{2-6} 
& Avg. $\lambda (\%)$ & $64.69$ & $10.32$ & $93.89$ & $809.8 \% $ \\ \hline
\end{tabular}
\caption{Average weight distribution variance and effective sample portion along the time axis. There is a sharp increase in the quality of H-iLQR before and after the hybrid event. The last column calculates the relative increase (resp. decrease) in the effective sample portion (resp. variance) before and after the jump, showing the influence of the hybrid events on feedback controllers for stochastic systems.}
\label{tab:var_lbd_avg_time}
\end{table*}

\subsection{Sample Efficiency and Proposal Controller Quality}
We are interested in studying the sampling process's variance and efficiency under the H-iLQR proposal controller. The variance of $\alpha$ and the effective sample portion, $\lambda$, are two metrics that measure the quality of H-iLQR-controlled hybrid path distribution. We show the statistics of these two indicators for all the $100$ experiments in $[0,T]$ in Fig. \ref{fig:bouncing_slip_var_lbd}. 

In Fig. \ref{fig:bouncing_slip_var_lbd}, we observe stronger vibrations in the variance and effective sample portion before the bounce time than after, for different noise realizations, for both SLIP and bouncing ball. As all the samples before the bounce time will experience the bouncing event, this indicates that the hybrid transition affects the robustness of the feedback controller used by these samples. Jump dynamics are the primary source of instabilities for hybrid systems. 

Nonlinear reset maps intensify this issue. Fig. \ref{fig:bouncing_lbd} shows the results for the bouncing ball dynamics with linear smooth flow and reset map. The $\lambda$ curve stays similar before and after the bounce. The main factor that affects the feedback controller is the mode mismatch brought by the uncertainties. By contrast, in Fig. \ref{fig:slip_lbd}, we see a drastic increase in $\lambda$ after the hybrid event. At event time, the system enters the linear flight dynamics from the nonlinear stance dynamics through a nonlinear reset map \eqref{eq:reset_slip_21}. 

Tab. \ref{tab:var_lbd_avg_time} separately computes the averages before and after the hybrid event. For the linear reset map in the bouncing ball example, the $\lambda$ increases from $65.83\%$ to $79.85\%$, and for the nonlinear SLIP model, $\lambda$ increases from $10.32\%$ to $93.89\%$.

\subsection{Ablation studies}
\label{sec:experiments_ablationstudy}

{\em a) Mode Mismatch and Trajectory Extensions.}

\begin{figure}[th]
    \centering
    \includegraphics[width=0.85\linewidth]{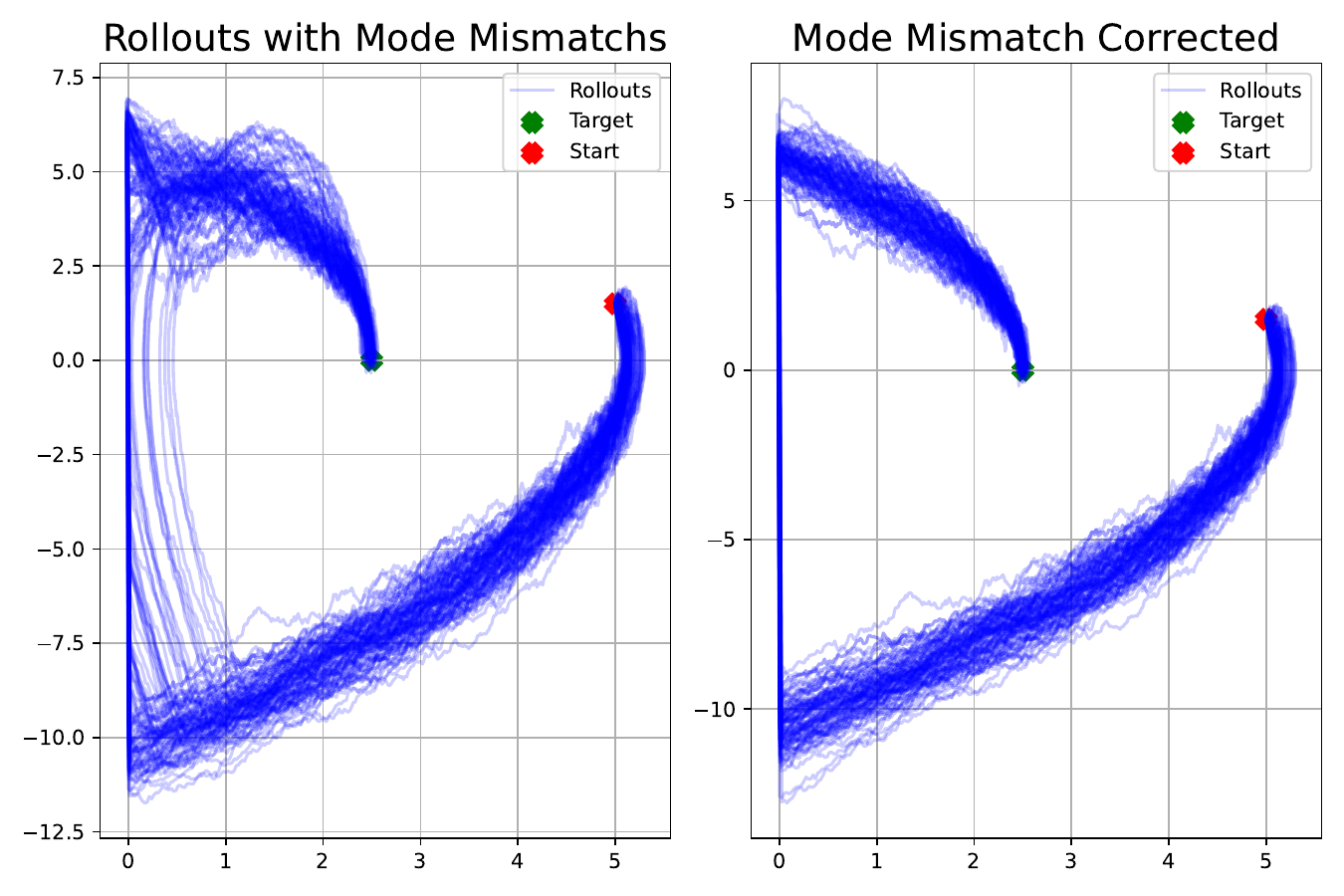}
    \caption{Stochastic rollout with mode mismatch (left) and the trajectory extension correction (right). }
    \label{fig:mode_mismatch_comparison}
\end{figure}

Mode mismatches affect the quality of the stochastic rollouts. Fig. \ref{fig:mode_mismatch_comparison} compares the trajectory rollouts at $t=0$, under the same randomness, with and without correcting mode mismatch. Visually, with the mode mismatches on the left-hand side, many sampled trajectories are dragged to the wrong modes without bouncing. After adding trajectory extensions, this dragging effect disappears in the right-hand side figure. 

We do an ablation study to investigate the improvement the mode mismatch correction brings to the path integral control rollouts. Tab. \ref{tab:comparison_modemismatch} records the variance and effective samples before and after the correction. Significant improvement in the sample efficiency is obtained. The effective sample portion reaching $64.63 \%$ shows that the H-iLQR controller with mode mismatch correction achieves near-optimal performance for the bouncing ball dynamics. 
\begin{table}[ht]
    \centering
    \begin{tabular}{|c|c|c|}
    \hline
         & Without Extensions & With Extensions \\
         \hline
       Variance              &  $49.98$    & $\textbf{0.55}$ \\
       \hline
       Effective Samples $(\%)$  & $1.96$ & $\textbf{64.63}$
       \\
       \hline
    \end{tabular}
    \caption{Comparison of variance and effective samples for path integral control updates, before and after mode mismatch correction.}
    \label{tab:comparison_modemismatch}
\end{table}

\begin{figure}[ht]
\centering
\begin{subfigure}{0.45\textwidth}
    \centering
    \includegraphics[width=0.6\linewidth]{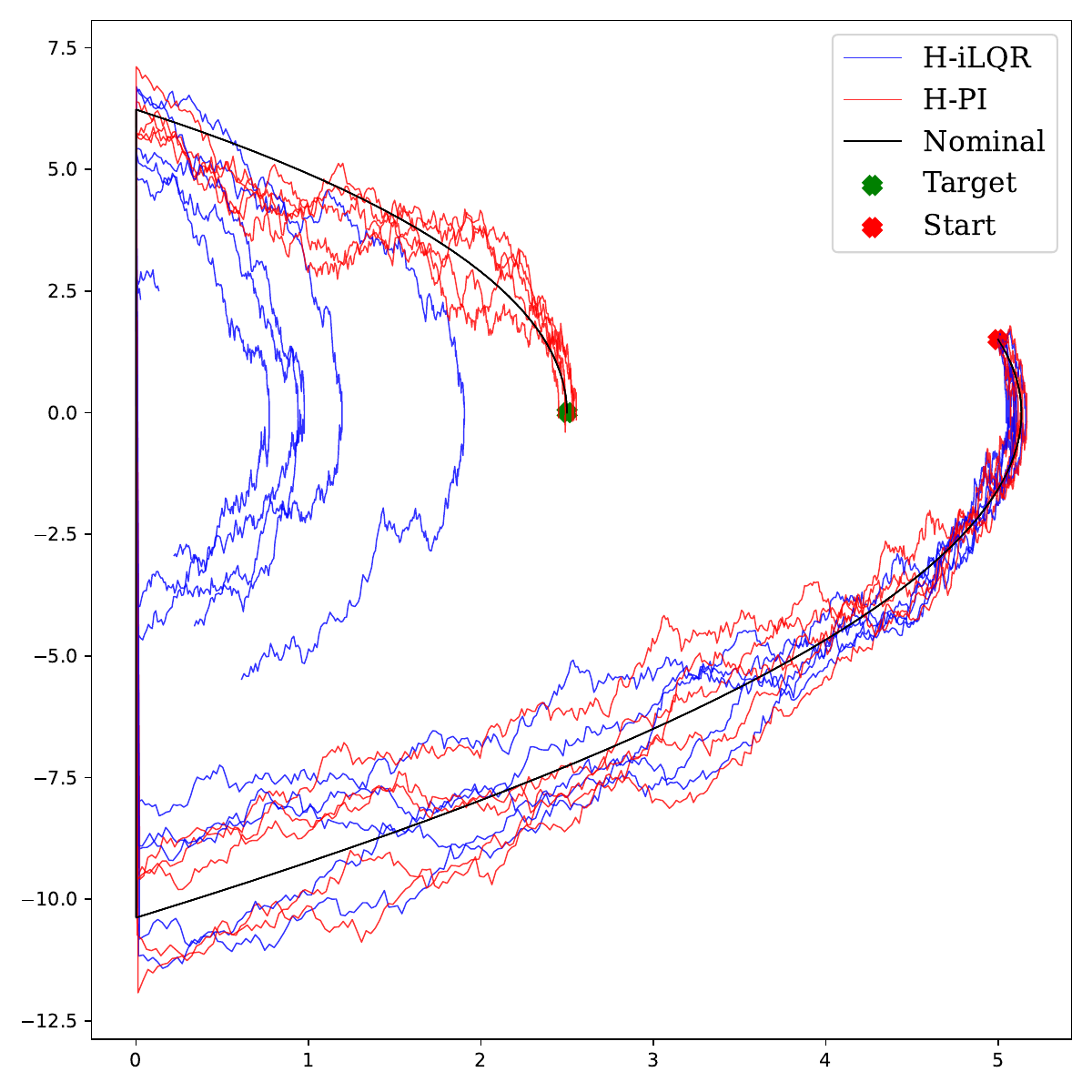}
    \caption{Obtaining optimal control from the zero-control rollouts. This figure shows the tail $5 \%$ costs H-iLQR controlled trajectories and the corresponding updated H-PI controlled trajectories.}
    \label{fig:bouncing_ball_tail_10_zerocontrol_1}
\end{subfigure}
\begin{subfigure}{0.235\textwidth}
    \includegraphics[width=\linewidth]{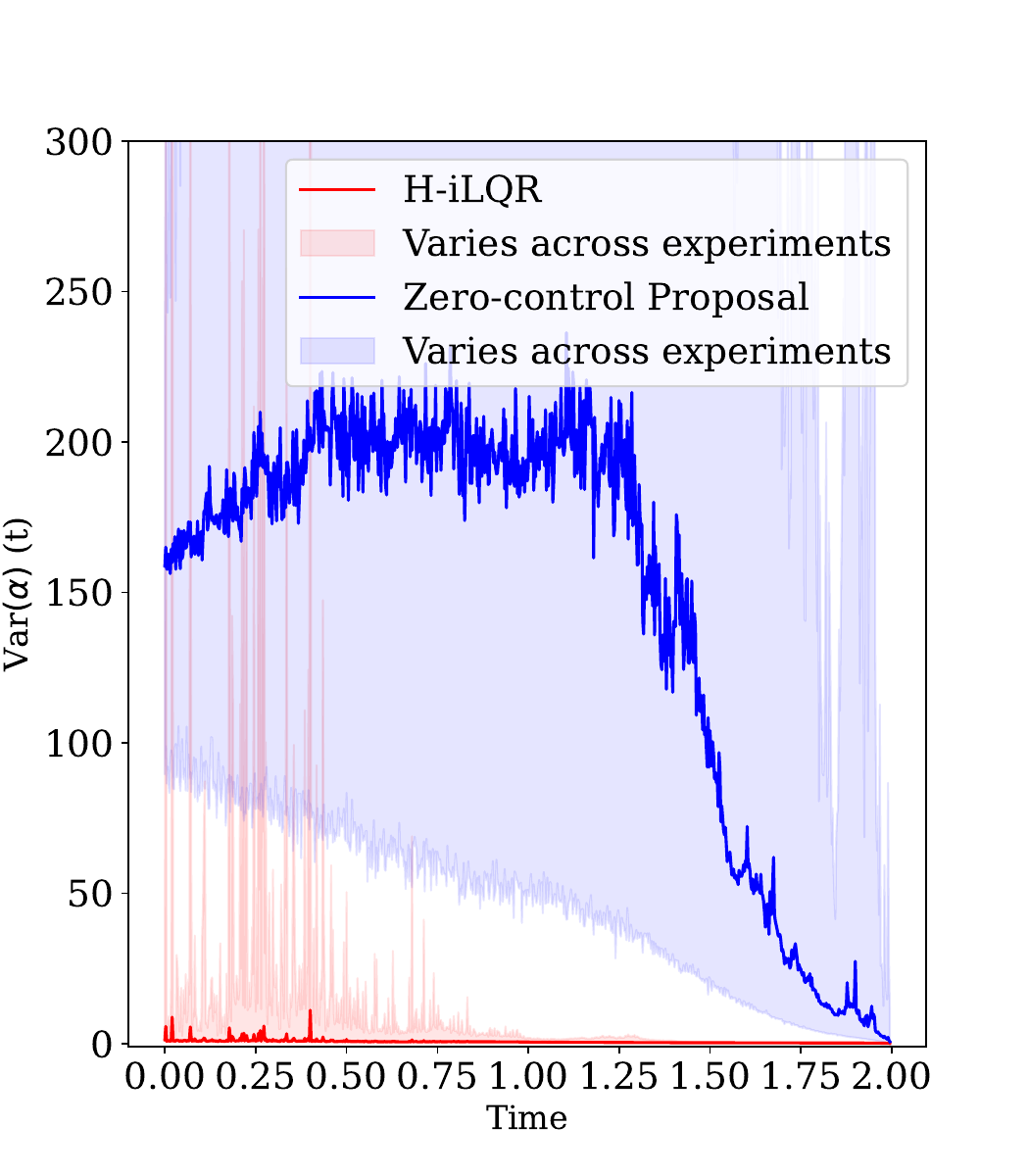}
    \caption{Variance Comparison.}
    \label{fig:bouncing_ball_tail_10_zerocontrol_2}
\end{subfigure}
\begin{subfigure}{0.24\textwidth}
    \includegraphics[width=\linewidth]{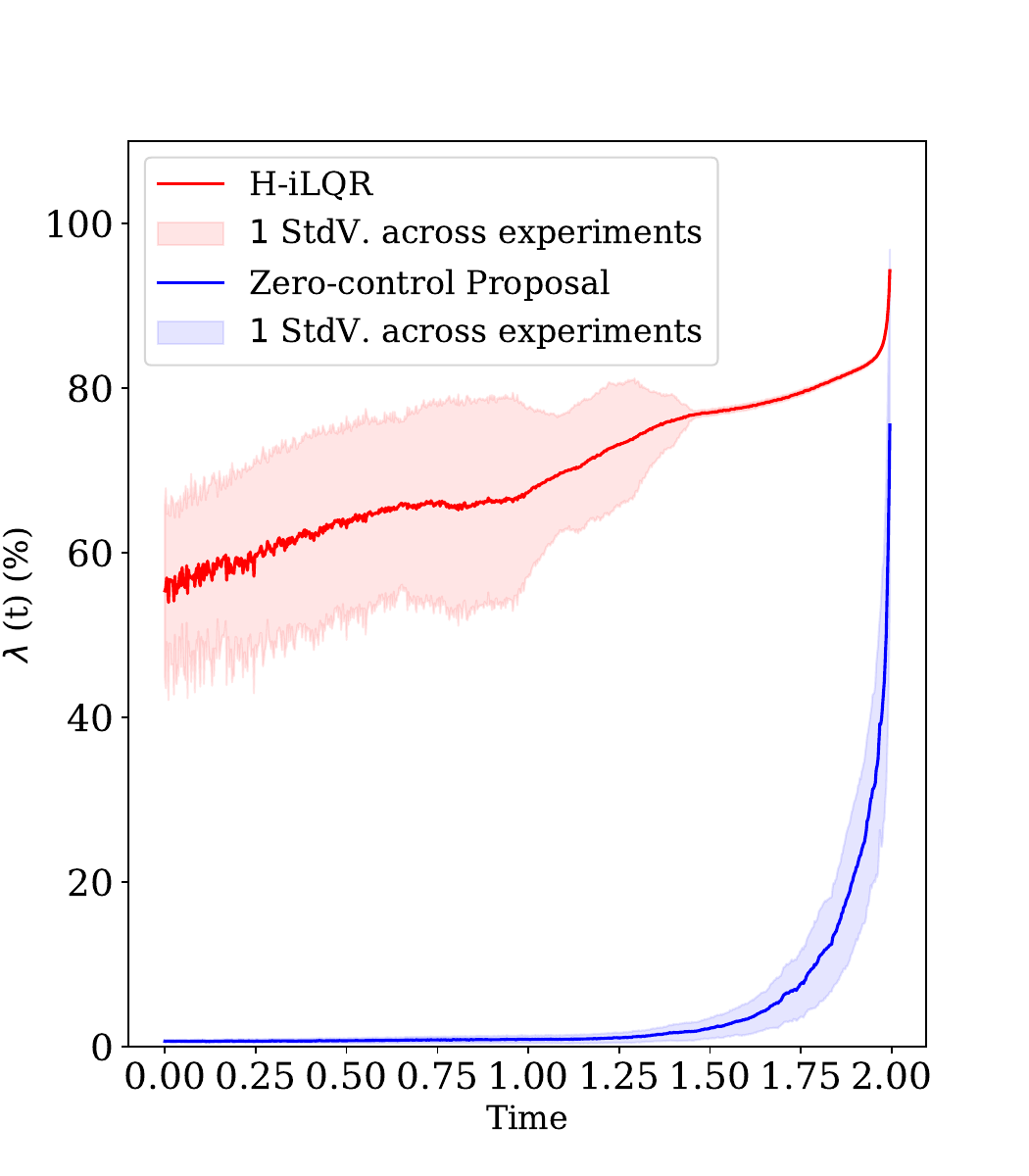}
    \caption{Effective Sample Comparison.}
    \label{fig:bouncing_ball_tail_10_zerocontrol_3}
\end{subfigure}
    \caption{Controlled trajectories under H-PI and the zero-control proposal are shown in subfigure \ref{fig:bouncing_ball_tail_10_zerocontrol_1}. Comparison of the variance and effective samples are shown in sub-figure \ref{fig:bouncing_ball_tail_10_zerocontrol_2} and \ref{fig:bouncing_ball_tail_10_zerocontrol_3}. H-iLQR provides a much lower variance proposal path distribution than $\mP^0$.}
    \label{fig:bouncing_ball_tail_10_zerocontrol}
\end{figure}

{\em (b) Importance sampling variance reduction through H-iLQR.} 
The optimal controller $u_t^*$ can be evaluated via a path integral over the hybrid path distributions induced by the uncontrolled process $\mP^0$, as in \eqref{eq:ustar_P0}. H-iLQR is a proposal controller in the importance sampling step to evaluate this path integral with reduced variance, as shown in Section \ref{sec:importance_sampling}. This ablation study uses the zero-controller, $u_t=0, \forall t$, to verify the optimal control law \eqref{eq:ustar_P0} and compare the variance with the H-iLQR induced path distribution in evaluating the path integral. Fig. \ref{fig:bouncing_ball_tail_10_zerocontrol} shows the controlled trajectories under the zero-control proposal and the improved results of H-PI for the tail $10\%$ experiments, and the quality of the zero-control proposal in the whole time window. Compared with the results in Fig. \ref{fig:bouncing_var} and Fig. \ref{fig:bouncing_lbd}, we can see that the H-iLQR proposal achieved much nearer regions to the optimal controller and thus reduced the sampling variance and complexity for H-PI.

\begin{table}[H]
    \centering
    \begin{tabular}{|c|c|c|c|}
    \hline
     & Zero Proposal  & H-PI & Reduced Cost $(\%)$
    \\ \hline
     Expected Cost & $2780.24$ & $59.56$ & $\textbf{97.86}$ 
    \\ \hline
    \end{tabular}
    \caption{Cost improvement over the zero-control proposal. }
\label{tab:bouncing_zerocontrol}
\end{table}

% {\em (d) Coupling in the randomness. }

\section{conclusion}
\label{sec:conclusion}
In this work, we solve optimal control problems for hybrid systems under uncertainties. We formulated the problem as a stochastic control problem with guard condition and reset function constraints. The ratio between two measures induced by SDEs with different drift terms and subject to guard conditions is given in this work, which has a similar form as the smooth stochastic systems. Leveraging the probability measure ratio, we then convert the hybrid stochastic control problem into a hybrid path distribution control problem, where the optimal controller can be obtained by forward sampling of SDEs with hybrid guard and reset constraints. This sampling process is parallelizable on GPU to efficiently solve the nonlinear stochastic control problem for hybrid systems. The sampling process is guided by a suboptimal Hybrid-iLQR controller to improve the sampling efficiency. Numerical experiments have been conducted to validate the proposed framework.

\appendix
\subsection{Proof of Lemma \ref{thm:change_of_measure}.}
\label{sec:appendix_proof_change_of_measure}
\begin{proof}
In each mode $I_j$, we write 
\[
\sqrt{\epsilon} \Delta W^j_k = X^j_{k+1} - (X^j_k + F_{j}(X^j_k))\Delta t,
\]
and the smooth conditional transition probability for \eqref{eq:nonlinear_SDE_1} is
\begin{align}
\label{eq:probability_p_Xk}
    & p(X^j_{k+1}|X^j_k) \nonumber
    \\
    & = Z \exp ( -\frac{\lVert X^j_{k+1} - (X^j_k + F_{j}(X^j_k)\Delta t) \rVert_{(\sigma_j \sigma_j')^{-1}}^2}{2\epsilon \Delta t} ) \nonumber
    \\
    &= Z \exp ( - \frac{1}{2 \epsilon \Delta t } \lVert \sqrt{\epsilon} \Delta W^j_k \rVert_{(\sigma_j \sigma_j')^{-1}}^2 ),
\end{align}
where $Z = \frac{1}{\sqrt{2\pi \epsilon \Delta t |(\sigma_j \sigma_j')^{-1}|}}$ is a normalizing constant. 

At jump time $k^{-}$, we have
\begin{equation}
\label{eq:jump_dyn_k}
    X^2_{k^{-}} = R_{12}(X^{1}_{k^{-}-1} + F_{1}(X^1_{k^{-}-1})\Delta t + \sqrt{\epsilon}\Delta W^1_{k^{-}-1}).
\end{equation}
The smooth dynamics \eqref{eq:nonlinear_SDE_2} is equivalently
\[
X^j_{k+1} = X^j_k + F_{j, k}\Delta t + \sqrt{\epsilon} \Delta W^j_k + (H_{j,k} - F_{j,k}) \Delta t,
\]
where $j\in\{1,2\}$ depending on the time $k$. The probability measure $d\mQ$ is independent of the trajectory, and we still have the jump at $k^-$. The transition probability under $d\mQ$ is
\begin{align}
\label{eq:probability_q_Xk}
\begin{split}
    \!\!\!\!\!\! & q(X^j_{k+1}|X^j_k) 
     \\
     & \!=\! Z \exp ( -\frac{\lVert \sqrt{\epsilon} \Delta W^j_k - (H_{j,k} - F_{j,k}) \Delta t \rVert_{(\sigma_j \sigma_j')^{-1}}^2}{2\epsilon \Delta t})
     \\
     & \!=\! Z \exp ( -\frac{\lVert \sqrt{\epsilon} \Delta W^j_k \rVert_{(\sigma_j \sigma_j')^{-1}}}{2\epsilon \Delta t} + \frac{(H_{j,k} - F_{j,k})' \Delta W^j_k }{\sqrt{\epsilon} \Delta t} 
     \\
     &\;\;\;\;\;\;\;\;\;\;\;\;\;\;\;\;\; - \frac{\lVert H_{j,k} - F_{j,k} \rVert_{(\sigma_j \sigma_j')^{-1}}^2 \Delta t}{2\epsilon})
\end{split}
\end{align}
Replace \eqref{eq:probability_q_Xk} into \eqref{eq:path_measure_mQ}, and replace \eqref{eq:probability_p_Xk} into \eqref{eq:path_measure_mP}, we have 
\begin{align*}
    \frac{d \mQ_H}{d\mP_H} 
    = & \exp \left[ \sum_{k=0}^{k^{-}-1} \left( \frac{(H_{1, k}-F_{1, k})' (\sigma_1 \sigma_1')^{-1} \Delta W^1_k}{\sqrt{\epsilon}}  \right. \right.
    \\
    & \left. \left. \;\;\;\; - \frac{\lVert H_{1, k}-F_{1, k} \rVert_{(\sigma_1 \sigma_1')^{-1}}^2\Delta t}{2\epsilon}  \right) \right.
    \\
    & \left. + \sum_{k=k^{-}}^{N_T} \left( \frac{(H_{2, k}-F_{2, k})' (\sigma_2 \sigma_2')^{-1} \Delta W^2_k}{\sqrt{\epsilon}} \right. \right. 
    \\ 
    & \left. \left.\;\;\;\;  - \frac{\lVert H_{2, k}-F_{2, k} \rVert_{(\sigma_2 \sigma_2')^{-1}}^2\Delta t}{2\epsilon}  \right) \right]
    \\
    = & \exp \left( \sum_{k=0}^{N_J} \int_{t_j^{+}}^{t_{j+1}^{-}} \frac{(H_{j}-F_{j})' (\sigma_j \sigma_j')^{-1} d W^j_t}{\sqrt{\epsilon}} \right. 
    \\
    & \left. \;\;\;\; - \frac{\lVert H_{j} - F_{j} \rVert_{(\sigma_j \sigma_j')^{-1}}^2 dt}{2\epsilon}  \right).
\end{align*}
Similarly, the ratio between the measure $d\mP$ and $d\mQ$ is 
\begin{align*}
\frac{d\mP_H}{d\mQ_H} = & \exp \left( \sum_{k=0}^{N_J} \int_{t_j^{+}}^{t_{j+1}^{-}} \frac{(F_{j}-H_{j})' (\sigma_j \sigma_j')^{-1} d W^j_t}{\sqrt{\epsilon}} \right. 
\\
& \left. \;\;\;\; + \frac{\lVert F_{j} - H_{j} \rVert_{(\sigma_j \sigma_j')^{-1}}^2 dt}{2\epsilon}  \right).
\end{align*}
\end{proof}

\subsection{Proof of Lemma \ref{thm:main_optimal_control}}
\label{sec:proof_optimal_control_expectation}
\begin{proof}
    Consider the definition of the optimal distribution $\mP^*$ in \eqref{eq:optimal_distribution_P_star}, the objective function in \eqref{eq:KL_P_optimal_Pu} is identical to 
    \begin{align*}
\label{eq:objective_optimal_control}
        \mE_{\mP^*} \! \left[\sum_{j=0}^{N_J}  \left (\int_{t^+_j}^{t^-_{j+1}} \frac{\lVert u^j_t \rVert^2}{2} dt - \sqrt{\epsilon} (u^j_t)' d W^j_t \right) \right],
    \end{align*}
    where $dW^j_t$ is a Wiener process under the measure $\mP^0$ in mode $I_j$.
    Consider our definition of the time discretizations \eqref{eq:time_discretization} and the hybrid discrete-time dynamics \eqref{eq:discretetime_smooth_map} and \eqref{eq:discretetime_jump_map}. For notation simplicity, the mode-dependent sequence 
    \begin{align*}
        \mathcal{U} &\triangleq \{ [u^0_0, \dots, u^0_{t^-_1}], \dots, [u^{N_J}_{t^+_{N_J}}, \dots, u^{N_J}_{t^-_{N_J+1}}] \}, \\
        \mathcal{W} &\triangleq \{ [\Delta W^0_0, \dots, \Delta W^0_{t^-_1}], \dots, [\Delta W^{N_J}_{t^+_{N_J}}, \dots, \Delta W^{N_J}_{t^-_{N_J+1}}] \}
    \end{align*}
    is written using the indexes in the discretization \eqref{eq:time_discretization} 
    \begin{equation*}
        \mathcal{U} = \{ u_0, \dots u_{N_T}\}, \; \mathcal{W} = \{ \Delta W_0, \dots, \Delta W_{N_T} \}.
    \end{equation*}
    The above controls $u_i$ have different physical meanings depending on time and mode. With this notation, the objective \eqref{eq:KL_P_optimal_Pu} is identically
    \begin{align*}
        &\mE_{\mP^*} \left[ \sum_{i=0}^{N_T} \frac{1}{2}\lVert u_i \rVert^2 \Delta t - \sqrt{\epsilon} u_i'\Delta W_i  \right]
        \\
        = & \sum_{i=0}^{N_T} \left(\frac{1}{2}\lVert u_i \rVert^2 \Delta t \right) - \sqrt{\epsilon} \sum_{i=0}^{N_T} u_i' \mE_{\mP^*} \left[ \Delta W_i \right].
    \end{align*}
    Set the gradient of the above at time $t$ to $0$, we have  
    \begin{equation*}
    \label{eq:ustar_expectation}
        u_t^* = \frac{\sqrt{\epsilon}}{\Delta t} \mE_{\mP^*} \left[ \Delta W_t \right].
    \end{equation*}
    The optimal control is obtained by evaluating the expectation over $\mP^*$, and $W_t$ is a Wiener process under the measure $\mP^0$. However, the distribution $\mP^*$ is impossible to sample from. We express $u_t^*$ as an expectation over $\mP^0$ by change of measure
    \[
    u_t^* = \frac{\sqrt{\epsilon}}{\Delta t} \mE_{\mP^0} \left[ \Delta W_t \frac{d\mP^*}{d\mP^0} \right].
    \]
    Plugging the definition of $\mP^*$ \eqref{eq:optimal_distribution_P_star}, we get the resulting \eqref{eq:ustar_P0}.
\end{proof}

\subsection{Backward Pass of H-iLQR}
\label{sec:appendix-hilqr}
For the deviated state and control $\delta X_i^j \triangleq X^j_i - \bar x_i, \; \delta u_i^j \triangleq u^j_i - \bar u_i,$ the function $Q$ can then be quadratically approximated, 
\begin{equation}
\label{eq:smooth_Q}
\!\!Q(\delta X^j_i, \delta u^j_i) \approx 
\frac{1}{2}
    \begin{bmatrix}
         1 \\
         \delta X^j_i
         \\
         \delta u^j_i
    \end{bmatrix}'
    \begin{bmatrix}
        0 & \partial_x Q' & \partial_u Q' 
        \\
        \partial_x Q & \partial_{xx} Q & \partial_{ux} Q'
        \\
        \partial_u Q & \partial_{ux} Q & \partial_{uu}Q
    \end{bmatrix}
    \begin{bmatrix}
         1 \\
         \delta X^j_i
         \\
         \delta u^j_i
    \end{bmatrix}.
\end{equation}
and the optimal increments to the approximated problem is
\[
(\delta X_i^*, \delta u_i^*) = \arg \min Q(\delta X_i^j, \delta u_i^j).
\]
Hybrid transition \eqref{eq:discretetime_jump_map} at $t_j^-$ can be approximated using the Saltation Matrix \eqref{eq:saltation_approximation_dyn} as
\begin{equation}
\label{eq:saltation_jump_map}
    \delta X^{j+1}_{t^{+}_j} \approx \Xi_{j,j+1}(t^{-}_{i}, t^{+}_{j}) \delta X^j_{t^{-}_{j}},
\end{equation} 
and the partial derivatives of the hybrid dynamics in $[t_i, t_{i+1}]$ is approximated to the first order by 
\begin{equation}
\label{eq:derv_f_hybrid}
    \partial_x F^{H}_{\Delta}(X_i, u_i) \approx \Xi_{j,j+1} \partial_x F_{\Delta, i}^j(X_i, u_i),
\end{equation}
leading the approximation of $Q$ at the jump time to be \cite{KongHybridiLQR}
\begin{equation*}
\begin{split}
    \partial_x Q^{H}_{i} &\coloneqq \partial_x l_{\Delta} + (\partial_x F_{\Delta, i}^j)' \Xi' (\partial_x V),
    \\
    \partial_u Q^{H}_{i} &\coloneqq \partial_u l_{\Delta} + (\partial_u F_{\Delta, i}^j)' \Xi' (\partial_x V),
    \\
    \partial_{xx}Q^{H}_{i} &\coloneqq \partial_{xx} l_{\Delta} + (\partial_x F_{\Delta, i}^j)' \Xi' (\partial_{xx}V) \Xi (\partial_x F_{\Delta, i}^j),
    \\
    \partial_{ux}Q^{H}_{i} &\coloneqq \partial_{ux} l_{\Delta} + (\partial_u F_{\Delta, i}^j)' \Xi' (\partial_{xx}V) \Xi (\partial_x F_{\Delta, i}^j),
    \\
    \partial_{uu} Q^{H}_{i} &\coloneqq \partial_{uu} l_{\Delta} + (\partial_u F_{\Delta, i}^j)' \Xi' (\partial_{xx}V) \Xi (\partial_u F_{\Delta, i}^j).
\end{split}
\end{equation*}
Minimizing the quadratic function \eqref{eq:smooth_Q} gives the optimal control increment
\begin{equation*}
    \delta u^*_i = - (\partial_{uu}Q_{i})^{-1}(\partial_u Q_{i}+ (\partial_{ux}Q_{i})\delta X_i) \triangleq - K_i\delta X_i + k_i
\end{equation*}
and the suboptimal controller to the original problem $u_t = \bar u_t + \delta u^*_i.$ This process is iterative until convergence.

\subsection{Mode mismatch and trajectory extensions}
\label{sec:method_mode_mismatch}
The reference trajectory extensions used in this work was introduced in \cite{rijnen2015optimal, KongHybridiLQR}. Specifically, at the $j{\rm th}$ hybrid transition, two extended trajectories can be computed over the time discretizations $[t^-_j, t_{1}, \dots, t_{N_{f}}], \; [t_{1}, \dots, t_{N_{b}}, t^+_{j+1}]$, starting from the pre- and post-contact state, respectively. The discrete forward reference extension is denoted as
\[
\{\bar{x}^{\rm fwd}\}_{N_f} \triangleq X(t^{-}_j) \cup \{\bar{x}^j_i\}_{i=1}^{N_f}, \forall i = 1,\dots, N_{f},
\]
where $\bar{x}^{j}_i = \int_{t^-_j}^{t_{i}} F_j(\tau, x^j_\tau)  + \sigma_{j}(\tau, x^j_\tau) \hat{u}^j d \tau + X^j_{t^-_j},$ and the backward extension is defined as
\[
\{\bar{x}^{\rm bwd}\}_{N_b} \triangleq X^{j+1}(t^{+}_j) \cup \{\bar{x}^{j+1}_i\}_{i=1}^{N_b}, \forall i = 1,\dots, N_{b}
\]
with
\begin{equation}
    \bar{x}^{j+1}_i \! = \! \int_{t^+_j}^{t_{i}} F_{j+1}(\tau, x^{j+1}_\tau) + \sigma_{j+1}(\tau, x^{j+1}_\tau) \hat{u}^{j+1} d \tau + X^{j+1}_{t^+_j}.
\end{equation}
The horizons $N_f, N_b$ are chosen to be long enough to cover all mismatches during the rollouts. The extensions can resolve the mode dimension mismatches. However, they are computed under some heuristic control $\hat{u}^j, \hat{u}^{j+1}$ \cite{rijnen2015optimal, KongHybridiLQR}. We also need to assign the feedback gains, $(\{\hat{K}^{\rm fwd}_i\}_{N_f}, \{\hat{K}^{\rm bwd}_i\}_{N_s})$, and feedforward gains $(\{\hat{k}^{\rm fwd}_i\}_{N_f}, \{\hat{k}^{\rm bwd}_i\}_{N_s})$ for the two extensions. In general, there is no optimal way to assign these gains. For instance, the terminal state in the forward extension is in mode $I_j$, while the target state $X_T$ may be in a different mode, making it impossible to compute the terminal loss using DP. In this work, we use the constant gains at $X^j_{t^-}$ and $X^{j+1}_{t^+}$ for the forward and backward extensions.

\begin{figure}[th]
\centering
\begin{subfigure}[b]{0.45\textwidth}
\centering
    \includegraphics[width=0.95\linewidth]{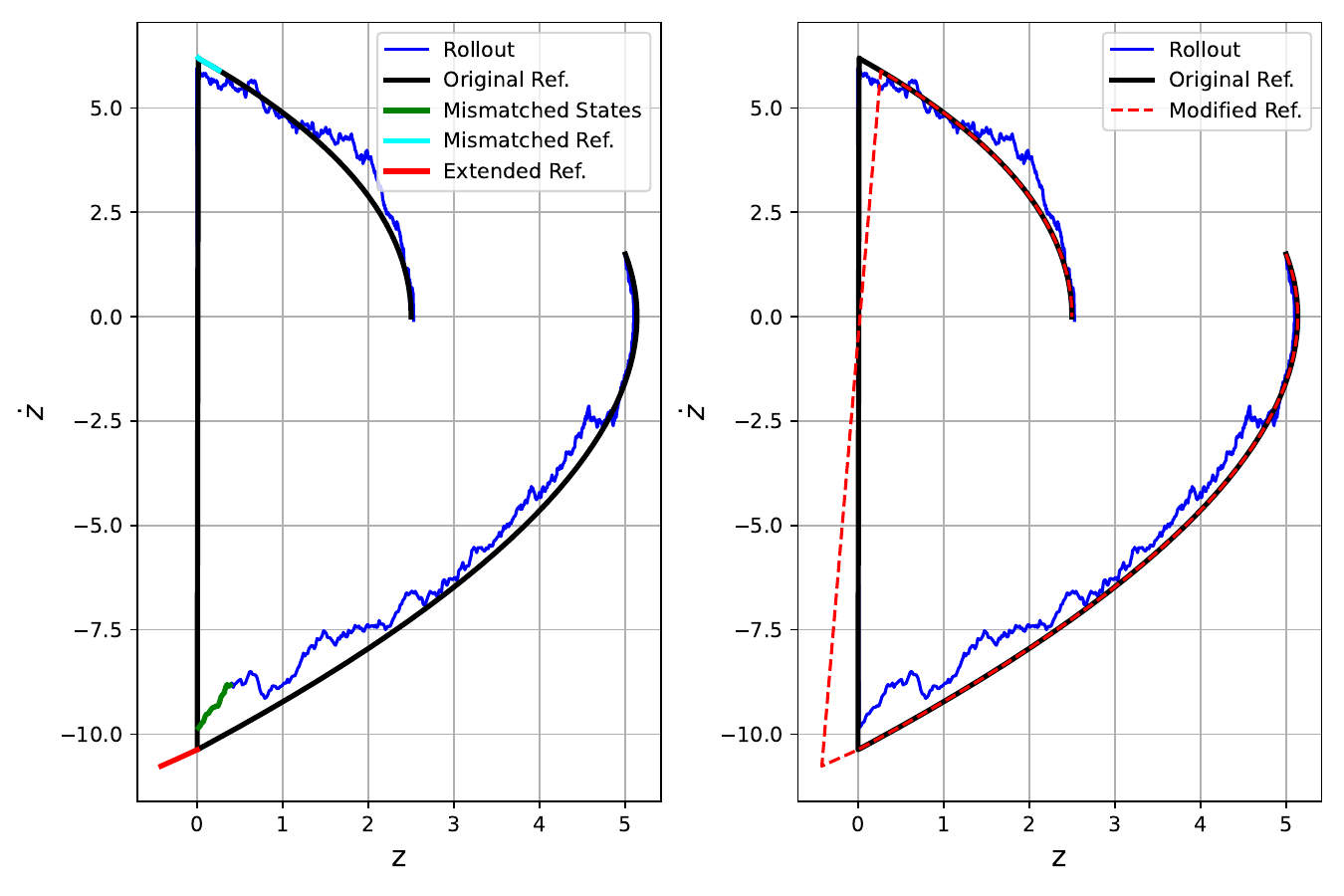}
    \caption{Late arrival and forward reference extension.}
    \label{fig:late_arrival}
\end{subfigure}
\begin{subfigure}[th]{0.45\textwidth}
    \centering
    \includegraphics[width=0.95\linewidth]{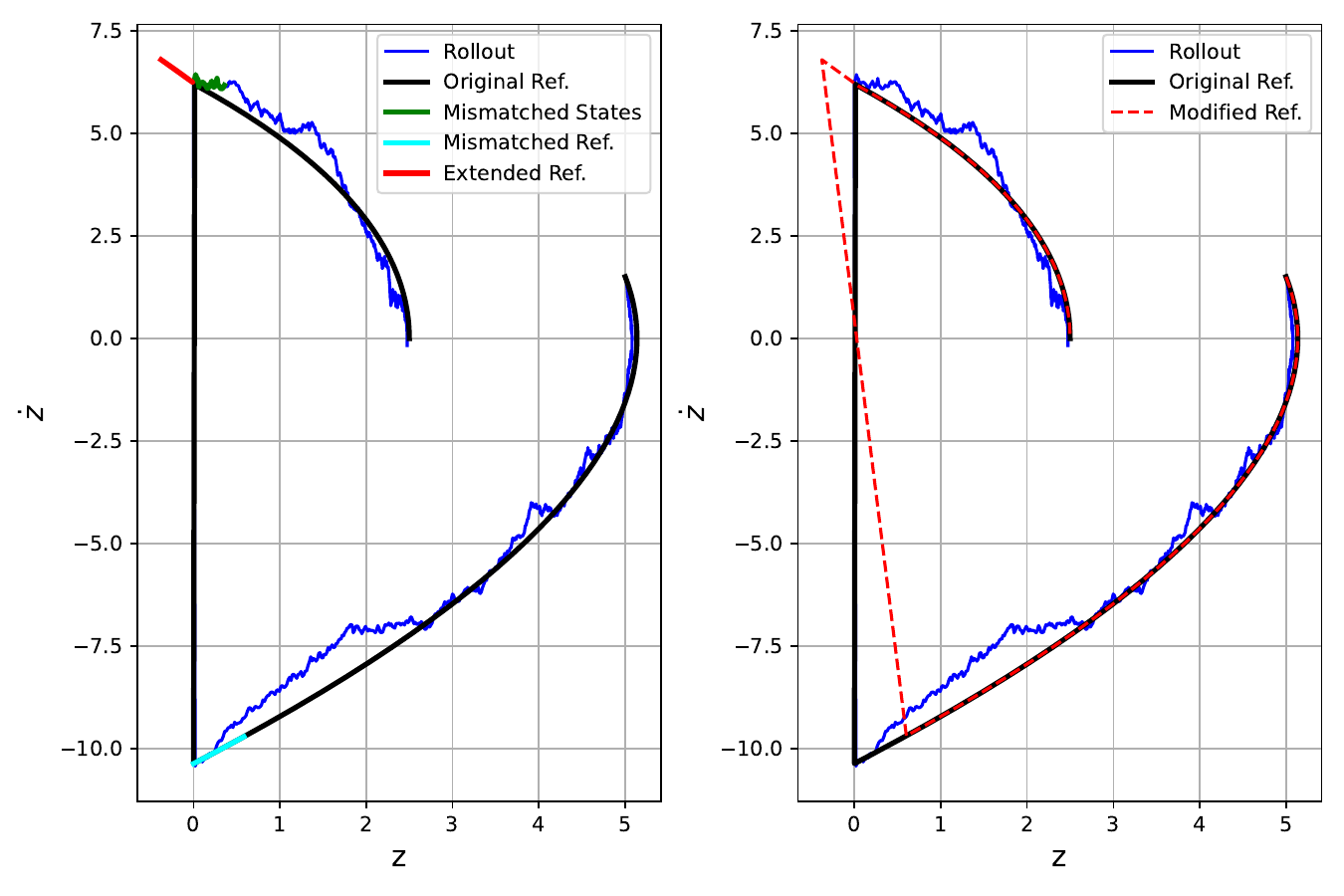}
    \caption{Early arrival and backward reference extension.}
    \label{fig:early_arrival}
\end{subfigure}
\caption{Mode mismatch and reference extensions during one rollout. Fig. \eqref{fig:late_arrival} shows the late arrival case, and Fig. \eqref{fig:early_arrival} shows the early arrival case. }
\label{fig:mode_mismatch}
\end{figure}

\bibliographystyle{Bibliography/IEEETrans}
\bibliography{root}

% Generated by IEEEtran.bst, version: 1.12 (2007/01/11)
\begin{thebibliography}{10}
\providecommand{\url}[1]{#1}
\csname url@samestyle\endcsname
\providecommand{\newblock}{\relax}
\providecommand{\bibinfo}[2]{#2}
\providecommand{\BIBentrySTDinterwordspacing}{\spaceskip=0pt\relax}
\providecommand{\BIBentryALTinterwordstretchfactor}{4}
\providecommand{\BIBentryALTinterwordspacing}{\spaceskip=\fontdimen2\font plus
\BIBentryALTinterwordstretchfactor\fontdimen3\font minus \fontdimen4\font\relax}
\providecommand{\BIBforeignlanguage}[2]{{%
\expandafter\ifx\csname l@#1\endcsname\relax
\typeout{** WARNING: IEEEtran.bst: No hyphenation pattern has been}%
\typeout{** loaded for the language `#1'. Using the pattern for}%
\typeout{** the default language instead.}%
\else
\language=\csname l@#1\endcsname
\fi
#2}}
\providecommand{\BIBdecl}{\relax}
\BIBdecl

\bibitem{thrun2002probabilistic}
S.~Thrun, ``Probabilistic robotics,'' \emph{Communications of the ACM}, vol.~45, no.~3, pp. 52--57, 2002.

\bibitem{maybeck1982stochastic}
P.~S. Maybeck, \emph{Stochastic models, estimation, and control}.\hskip 1em plus 0.5em minus 0.4em\relax Academic press, 1982.

\bibitem{aastrom2012introduction}
K.~J. {\AA}str{\"o}m, \emph{Introduction to stochastic control theory}.\hskip 1em plus 0.5em minus 0.4em\relax Courier Corporation, 2012.

\bibitem{bertsekas1996stochastic}
D.~Bertsekas and S.~E. Shreve, \emph{Stochastic optimal control: the discrete-time case}.\hskip 1em plus 0.5em minus 0.4em\relax Athena Scientific, 1996, vol.~5.

\bibitem{kushner1990numerical}
H.~J. Kushner, ``Numerical methods for stochastic control problems in continuous time,'' \emph{SIAM Journal on Control and Optimization}, vol.~28, no.~5, pp. 999--1048, 1990.

\bibitem{kappen2005path}
H.~J. Kappen, ``Path integrals and symmetry breaking for optimal control theory,'' \emph{Journal of statistical mechanics: theory and experiment}, vol. 2005, no.~11, p. P11011, 2005.

\bibitem{todorov2005generalized}
E.~Todorov and W.~Li, ``A generalized iterative lqg method for locally-optimal feedback control of constrained nonlinear stochastic systems,'' in \emph{IEEE American Control Conference (ACC)}, 2005, pp. 300--306.

\bibitem{chen2015optimal_1}
Y.~Chen, T.~T. Georgiou, and M.~Pavon, ``Optimal steering of a linear stochastic system to a final probability distribution, part i,'' \emph{IEEE Transactions on Automatic Control}, vol.~61, no.~5, pp. 1158--1169, 2015.

\bibitem{collins2005bipedal}
S.~H. Collins and A.~Ruina, ``A bipedal walking robot with efficient and human-like gait,'' in \emph{IEEE International Conference on Robotics and Automation (ICRA)}, 2005, pp. 1983--1988.

\bibitem{laszlo1996limit}
J.~Laszlo, M.~van~de Panne, and E.~Fiume, ``Limit cycle control and its application to the animation of balancing and walking,'' in \emph{Proceedings of the 23rd annual conference on Computer graphics and interactive techniques}, 1996, pp. 155--162.

\bibitem{todd2013walking}
D.~J. Todd, \emph{Walking machines: an introduction to legged robots}.\hskip 1em plus 0.5em minus 0.4em\relax Springer Science \& Business Media, 2013.

\bibitem{kuindersma2016optimization}
S.~Kuindersma, R.~Deits, M.~Fallon, A.~Valenzuela, H.~Dai, F.~Permenter, T.~Koolen, P.~Marion, and R.~Tedrake, ``Optimization-based locomotion planning, estimation, and control design for the atlas humanoid robot,'' \emph{Autonomous robots}, vol.~40, pp. 429--455, 2016.

\bibitem{clark2001biomimetic}
J.~E. Clark, J.~G. Cham, S.~A. Bailey, E.~M. Froehlich, P.~K. Nahata, R.~J. Full, and M.~R. Cutkosky, ``Biomimetic design and fabrication of a hexapedal running robot,'' in \emph{IEEE International Conference on Robotics and Automation (ICRA)}, vol.~4, 2001, pp. 3643--3649.

\bibitem{hutter2011scarleth}
M.~Hutter, C.~D. Remy, M.~A. Hoepflinger, and R.~Siegwart, ``Scarleth: Design and control of a planar running robot,'' in \emph{IEEE/RSJ international conference on intelligent robots and systems (IROS)}, 2011, pp. 562--567.

\bibitem{westervelt2018feedback}
E.~R. Westervelt, J.~W. Grizzle, C.~Chevallereau, J.~H. Choi, and B.~Morris, \emph{Feedback control of dynamic bipedal robot locomotion}.\hskip 1em plus 0.5em minus 0.4em\relax CRC press, 2018.

\bibitem{johnson2016hybrid}
A.~M. Johnson, S.~A. Burden, and D.~E. Koditschek, ``A hybrid systems model for simple manipulation and self-manipulation systems,'' \emph{The International Journal of Robotics Research}, vol.~35, no.~11, pp. 1354--1392, 2016.

\bibitem{billard2019trends}
A.~Billard and D.~Kragic, ``Trends and challenges in robot manipulation,'' \emph{Science}, vol. 364, no. 6446, p. eaat8414, 2019.

\bibitem{grossman1993hybrid}
R.~L. Grossman, A.~Nerode, A.~P. Ravn, and H.~Rischel, \emph{Hybrid systems}.\hskip 1em plus 0.5em minus 0.4em\relax Springer, 1993, vol. 736.

\bibitem{posa2014direct}
M.~Posa, C.~Cantu, and R.~Tedrake, ``A direct method for trajectory optimization of rigid bodies through contact,'' \emph{The International Journal of Robotics Research}, vol.~33, no.~1, pp. 69--81, 2014.

\bibitem{mordatch2012discovery}
I.~Mordatch, E.~Todorov, and Z.~Popovi{\'c}, ``Discovery of complex behaviors through contact-invariant optimization,'' \emph{ACM Transactions on Graphics (ToG)}, vol.~31, no.~4, pp. 1--8, 2012.

\bibitem{diehl2006fast}
M.~Diehl, H.~G. Bock, H.~Diedam, and P.-B. Wieber, ``Fast direct multiple shooting algorithms for optimal robot control,'' \emph{Fast motions in biomechanics and robotics: optimization and feedback control}, pp. 65--93, 2006.

\bibitem{manchester2011regions}
I.~R. Manchester, M.~M. Tobenkin, M.~Levashov, and R.~Tedrake, ``Regions of attraction for hybrid limit cycles of walking robots,'' \emph{IFAC Proceedings Volumes}, vol.~44, no.~1, pp. 5801--5806, 2011.

\bibitem{manchester2011stable}
I.~R. Manchester, U.~Mettin, F.~Iida, and R.~Tedrake, ``Stable dynamic walking over uneven terrain,'' \emph{The International Journal of Robotics Research}, vol.~30, no.~3, pp. 265--279, 2011.

\bibitem{yu2024optimal}
H.~Yu, D.~F. Franco, A.~M. Johnson, and Y.~Chen, ``Optimal covariance steering of linear stochastic systems with hybrid transitions,'' \emph{arXiv preprint arXiv:2410.13222}, 2024.

\bibitem{rijnen2015optimal}
M.~Rijnen, A.~Saccon, and H.~Nijmeijer, ``On optimal trajectory tracking for mechanical systems with unilateral constraints,'' in \emph{IEEE Conference on Decision and Control (CDC)}, 2015, pp. 2561--2566.

\bibitem{rijnen2017control}
M.~Rijnen, E.~De~Mooij, S.~Traversaro, F.~Nori, N.~Van De~Wouw, A.~Saccon, and H.~Nijmeijer, ``Control of humanoid robot motions with impacts: Numerical experiments with reference spreading control,'' in \emph{IEEE International Conference on Robotics and Automation (ICRA)}, 2017, pp. 4102--4107.

\bibitem{li2004iterative}
W.~Li and E.~Todorov, ``Iterative linear quadratic regulator design for nonlinear biological movement systems,'' in \emph{First International Conference on Informatics in Control, Automation and Robotics}, vol.~2.\hskip 1em plus 0.5em minus 0.4em\relax SciTePress, 2004, pp. 222--229.

\bibitem{KongHybridiLQR}
N.~J. Kong, C.~Li, G.~Council, and A.~M. Johnson, ``Hybrid ilqr model predictive control for contact implicit stabilization on legged robots,'' \emph{IEEE Transactions on Robotics}, vol.~39, no.~6, pp. 4712--4727, 2023.

\bibitem{kong2023saltation}
N.~J. Kong, J.~J. Payne, J.~Zhu, and A.~M. Johnson, ``Saltation matrices: The essential tool for linearizing hybrid dynamical systems,'' \emph{Proceedings of the IEEE}, 2024.

\bibitem{bertsekas2012dynamic}
D.~Bertsekas, \emph{Dynamic programming and optimal control: Volume I}.\hskip 1em plus 0.5em minus 0.4em\relax Athena scientific, 2012, vol.~4.

\bibitem{kappen2005linear}
H.~J. Kappen, ``Linear theory for control of nonlinear stochastic systems,'' \emph{Physical review letters}, vol.~95, no.~20, p. 200201, 2005.

\bibitem{GradyWilliamsICRA}
G.~Williams, N.~Wagener, B.~Goldfain, P.~Drews, J.~M. Rehg, B.~Boots, and E.~A. Theodorou, ``Information theoretic mpc for model-based reinforcement learning,'' in \emph{IEEE International Conference on Robotics and Automation (ICRA)}, 2017, pp. 1714--1721.

\bibitem{GradyWilliamsTRO}
G.~Williams, P.~Drews, B.~Goldfain, J.~M. Rehg, and E.~A. Theodorou, ``Information-theoretic model predictive control: Theory and applications to autonomous driving,'' \emph{IEEE Transactions on Robotics}, vol.~34, no.~6, pp. 1603--1622, 2018.

\bibitem{grizzle2014models}
J.~W. Grizzle, C.~Chevallereau, R.~W. Sinnet, and A.~D. Ames, ``Models, feedback control, and open problems of 3d bipedal robotic walking,'' \emph{Automatica}, vol.~50, no.~8, pp. 1955--1988, 2014.

\bibitem{zhangpath}
Q.~Zhang and Y.~Chen, ``Path integral sampler: A stochastic control approach for sampling,'' in \emph{International Conference on Learning Representations}.

\bibitem{zhang2023optimal}
Q.~Zhang, A.~Taghvaei, and Y.~Chen, ``An optimal control approach to particle filtering,'' \emph{Automatica}, vol. 151, p. 110894, 2023.

\bibitem{kappen2007introduction}
H.~J. Kappen, ``An introduction to stochastic control theory, path integrals and reinforcement learning,'' in \emph{AIP conference proceedings}, vol. 887, no.~1.\hskip 1em plus 0.5em minus 0.4em\relax American Institute of Physics, 2007, pp. 149--181.

\bibitem{del2004feynman}
P.~Del~Moral and P.~Del~Moral, \emph{Feynman-kac formulae}.\hskip 1em plus 0.5em minus 0.4em\relax Springer, 2004.

\bibitem{theodorou2010generalized}
E.~Theodorou, J.~Buchli, and S.~Schaal, ``A generalized path integral control approach to reinforcement learning,'' \emph{The Journal of Machine Learning Research}, vol.~11, pp. 3137--3181, 2010.

\bibitem{chen2015optimal_2}
Y.~Chen, T.~T. Georgiou, and M.~Pavon, ``Optimal steering of a linear stochastic system to a final probability distribution, part ii,'' \emph{IEEE Transactions on Automatic Control}, vol.~61, no.~5, pp. 1170--1180, 2015.

\bibitem{yongxin2018optimal_3}
------, ``Optimal steering of a linear stochastic system to a final probability distribution, part iii,'' \emph{IEEE Transactions on Automatic Control}, vol.~63, no.~9, pp. 3112--3118, 2018.

\bibitem{kappen2012optimal}
H.~J. Kappen, V.~G{\'o}mez, and M.~Opper, ``Optimal control as a graphical model inference problem,'' \emph{Machine learning}, vol.~87, pp. 159--182, 2012.

\bibitem{rawlik2013stochastic}
K.~Rawlik, M.~Toussaint, and S.~Vijayakumar, ``On stochastic optimal control and reinforcement learning by approximate inference,'' in \emph{Proceedings of the International joint Conference on Artificial Intelligence}, 2013, pp. 3052--3056.

\bibitem{zhang2014applications}
W.~Zhang, H.~Wang, C.~Hartmann, M.~Weber, and C.~Schütte, ``Applications of the cross-entropy method to importance sampling and optimal control of diffusions,'' \emph{SIAM Journal on Scientific Computing}, vol.~36, no.~6, pp. A2654--A2672, 2014.

\bibitem{yu2023gaussian}
H.~Yu and Y.~Chen, ``A gaussian variational inference approach to motion planning,'' \emph{IEEE Robotics and Automation Letters}, vol.~8, no.~5, pp. 2518--2525, 2023.

\bibitem{yu2023stochastic}
------, ``Stochastic motion planning as gaussian variational inference: Theory and algorithms,'' \emph{arXiv preprint arXiv:2308.14985}, 2023.

\bibitem{power2024constrained}
T.~Power and D.~Berenson, ``Constrained stein variational trajectory optimization,'' \emph{IEEE Transactions on Robotics}, 2024.

\bibitem{mangasarian1993nonlinear}
O.~L. Mangasarian and M.~V. Solodov, ``Nonlinear complementarity as unconstrained and constrained minimization,'' \emph{Mathematical Programming}, vol.~62, no.~1, pp. 277--297, 1993.

\bibitem{hager2009nonlinear}
C.~Hager and B.~Wohlmuth, ``Nonlinear complementarity functions for plasticity problems with frictional contact,'' \emph{Computer Methods in Applied Mechanics and Engineering}, vol. 198, no. 41-44, pp. 3411--3427, 2009.

\bibitem{manchester2019contact}
Z.~Manchester, N.~Doshi, R.~J. Wood, and S.~Kuindersma, ``Contact-implicit trajectory optimization using variational integrators,'' \emph{The International Journal of Robotics Research}, vol.~38, no. 12-13, pp. 1463--1476, 2019.

\bibitem{patel2019contact}
A.~Patel, S.~L. Shield, S.~Kazi, A.~M. Johnson, and L.~T. Biegler, ``Contact-implicit trajectory optimization using orthogonal collocation,'' \emph{IEEE Robotics and Automation Letters}, vol.~4, no.~2, pp. 2242--2249, 2019.

\bibitem{filippov2013differential}
A.~F. Filippov, \emph{Differential equations with discontinuous righthand sides: control systems}.\hskip 1em plus 0.5em minus 0.4em\relax Springer Science \& Business Media, 2013, vol.~18.

\bibitem{munoz2019enhancing}
J.-G. Mu{\~n}oz, A.~P{\'e}rez, and F.~Angulo, ``Enhancing the stability of the switched systems using the saltation matrix,'' \emph{International Journal of Structural Stability and Dynamics}, vol.~19, no.~05, p. 1941004, 2019.

\bibitem{belter2011rough}
D.~Belter and P.~Skrzypczy{\'n}ski, ``Rough terrain mapping and classification for foothold selection in a walking robot,'' \emph{Journal of Field Robotics}, vol.~28, no.~4, pp. 497--528, 2011.

\bibitem{tassa2011stochastic}
Y.~Tassa, ``Stochastic complementarity for local control of discontinuous dynamics,'' \emph{Robotics: Science and Systems VI}, p. 169, 2011.

\bibitem{dai2012optimizing}
H.~Dai and R.~Tedrake, ``Optimizing robust limit cycles for legged locomotion on unknown terrain,'' in \emph{IEEE Conference on Decision and Control (CDC)}, 2012, pp. 1207--1213.

\bibitem{shirai2024chance}
Y.~Shirai, D.~K. Jha, A.~U. Raghunathan, and D.~Romeres, ``Chance-constrained optimization for contact-rich systems using mixed integer programming,'' \emph{Nonlinear Analysis: Hybrid Systems}, vol.~52, p. 101466, 2024.

\bibitem{drnach2021robust}
L.~Drnach and Y.~Zhao, ``Robust trajectory optimization over uncertain terrain with stochastic complementarity,'' \emph{IEEE Robotics and Automation Letters}, vol.~6, no.~2, pp. 1168--1175, 2021.

\bibitem{thijssen2015path}
S.~Thijssen and H.~Kappen, ``Path integral control and state-dependent feedback,'' \emph{Physical Review E}, vol.~91, no.~3, p. 032104, 2015.

\bibitem{Gir60}
I.~V. Girsanov, ``On transforming a certain class of stochastic processes by absolutely continuous substitution of measures,'' \emph{Theory of Probability \& Its Applications}, vol.~5, no.~3, pp. 285--301, 1960.

\bibitem{aizerman1958stability}
M.~Aizerman and F.~Gantmakher, ``On the stability of periodic motions,'' \emph{Journal of Applied Mathematics and Mechanics}, vol.~22, no.~6, pp. 1065--1078, 1958.

\bibitem{kong2021ilqr}
N.~J. Kong, G.~Council, and A.~M. Johnson, ``ilqr for piecewise-smooth hybrid dynamical systems,'' in \emph{IEEE Conference on Decision and Control (CDC)}, 2021, pp. 5374--5381.

\bibitem{sarkka2019applied}
S.~S{\"a}rkk{\"a} and A.~Solin, \emph{Applied stochastic differential equations}.\hskip 1em plus 0.5em minus 0.4em\relax Cambridge University Press, 2019, vol.~10.

\end{thebibliography}

\end{document}